\documentclass[10pt,journal,compsoc,twoside]{IEEEtran}

\usepackage{amsmath,amssymb,amsfonts,bm}
\usepackage[colorlinks, linkcolor=magenta, anchorcolor=green, citecolor=blue, urlcolor=black]{hyperref}
\usepackage{xcolor}
\usepackage{algorithmic}
\usepackage{caption}
\usepackage{array}
\usepackage{textcomp}
\usepackage{stfloats}
\usepackage{url}
\usepackage{verbatim}
\usepackage{algorithmic}
\usepackage[linesnumbered,ruled]{algorithm2e}

\newcommand*{\circled}[1]{\lower.7ex\hbox{\tikz\draw (0pt, 0pt)%
    circle (.5em) node {\makebox[1em][c]{\small #1}};}}
\usepackage[square, numbers,sort&compress]{natbib}
\usepackage{bigstrut,multirow,rotating}
\usepackage{booktabs}
\usepackage{multirow}
\usepackage{array}
\usepackage{multirow}
\usepackage{rotating}
\usepackage{graphicx,pifont}
\usepackage[normalem]{ulem}

\let\oldding\ding
\renewcommand{\ding}[2][1]{\scalebox{#1}{\oldding{#2}}}
\def\BibTeX{{\rm B\kern-.05em{\sc i\kern-.025em b}\kern-.08em
    T\kern-.1667em\lower.7ex\hbox{E}\kern-.125emX}}

\begin{document}

\title{HGNAS: \underline{H}ardware-Aware \underline{G}raph \underline{N}eural \underline{A}rchitecture \underline{S}earch for Edge Devices}

\author{Ao~Zhou,~Jianlei~Yang,~\IEEEmembership{Senior~Member,~IEEE},~Yingjie Qi,~Tong~Qiao,~Yumeng Shi,\\
Cenlin Duan,~Weisheng~Zhao,~\IEEEmembership{Fellow,~IEEE}, Chunming Hu
\IEEEcompsocitemizethanks{
\IEEEcompsocthanksitem This work is supported in part by the National Natural Science Foundation of China (Grant No. 62072019) and the Fundamental Research Funds for the Central Universities.
\textit{Corresponding authors are Jianlei Yang and Chunming Hu.}
\IEEEcompsocthanksitem A. Zhou, J. Yang, Y. Qi, T. Qiao, Y. Shi and C. Hu are with School of Computer Science and Engineering, Beihang University, Beijing, China. Email: \url{jianlei@buaa.edu.cn}, \url{hucm@buaa.edu.cn}.
\IEEEcompsocthanksitem C. Duan and W. Zhao are with School of Integrated Circuits and Engineering, Beihang University, Beijing, China.
}
\thanks{Manuscript received on Sept 14, 2023, revised on May 28 and Aug 8, 2024, accepted on Aug 15, 2024.}
}

\IEEEtitleabstractindextext{
\begin{abstract}

Graph Neural Networks (GNNs) are becoming increasingly popular for graph-based learning tasks such as point cloud processing due to their state-of-the-art (SOTA) performance.
Nevertheless, the research community has primarily focused on improving model expressiveness, lacking consideration of how to design efficient GNN models for edge scenarios with real-time requirements and limited resources.
Examining existing GNN models reveals varied execution across platforms and frequent Out-Of-Memory (OOM) problems, highlighting the need for hardware-aware GNN design.
To address this challenge, this work proposes a novel hardware-aware graph neural architecture search framework tailored for resource constraint edge devices, namely HGNAS.
To achieve hardware awareness, HGNAS integrates an efficient GNN hardware performance predictor that evaluates the latency and peak memory usage of GNNs in milliseconds.
Meanwhile, we study GNN memory usage during inference and offer a peak memory estimation method, enhancing the robustness of architecture evaluations when combined with predictor outcomes.
Furthermore, HGNAS constructs a fine-grained design space to enable the exploration of extreme performance architectures by decoupling the GNN paradigm.
In addition, the multi-stage hierarchical search strategy is leveraged to facilitate the navigation of huge candidates, which can reduce the single search time to a few GPU hours.
To the best of our knowledge, HGNAS is the first automated GNN design framework for edge devices, and also the first work to achieve hardware awareness of GNNs across different platforms.
Extensive experiments across various applications and edge devices have proven the superiority of HGNAS. It can achieve up to a $10.6\times$ speedup and an $82.5\%$ peak memory reduction with negligible accuracy loss compared to DGCNN on ModelNet40.

\end{abstract}

}

\maketitle

\IEEEraisesectionheading{\section{Introduction}\label{sec:introduction}}

\IEEEPARstart{G}{raph} neural networks (GNNs) have been deemed as a promising engine of artificial intelligence (AI), achieving state-of-the-art (SOTA) performance in a wide range of real-world applications, such as node classification \cite{kipf2017semi}, link prediction \cite{zhang2018link}, recommendation system \cite{ying2018graph} and 3D representation learning \cite{qi2017pointnet++}. 
Due to the powerful feature extraction capabilities on topological structures, GNN has become a popular strategy for handling point cloud data\cite{li2021towards}, which reveals the prospect of GNN application in edge scenarios.
Moreover, with the increasing popularity of 3D scanning sensors in edge devices, such as mobile phones and unmanned aerial vehicles, it is an inevitable trend to deploy GNNs on various edge devices to embrace hardware intelligence \cite{shao2021branchy, zhou2024graph, qiao2024gnnavigator}.
However, the inference process of GNNs involves both compute-intensive and memory-intensive stages, resulting in a huge computational gap between resource-limited edge devices and expensive GNNs.
We deployed the popular Dynamic Graph Convolutional Neural Network (DGCNN) \cite{wang2019dynamic} used in point cloud processing on a Raspberry Pi, where it takes over 4 seconds to process a single frame and encounters Out-Of-Memory (OOM) problems when handling graphs with more than 1536 points \cite{zhou2023hardware}.
Therefore, deploying GNNs on edge devices for real-time inference is extremely challenging due to resource constraints.

For the purpose of tackling the prohibitive inference cost, several handcrafted approaches have been dedicated to designing resource-efficient GNNs for point cloud processing \cite{li2021towards, tailor2021towards}. 
However, given the expansive design space and diverse hardware characteristics, manual optimization incurs significant computational overhead due to the extensive trial-and-error required to identify layers for optimization.
In practice, manual optimization performance highly depends on human experience.
To minimize manual labor and address unstable on-device performance, hardware-aware neural architecture search (NAS) has recently emerged as a promising automated technique for customizing optimal CNNs for various application scenarios~\cite{dudziak2020brp, jiang2019accuracy}.
Inspired by these efforts, we seek to take the point cloud processing application as an opportunity to explore the prospect of hardware-aware NAS for designing efficient GNNs on edge devices.
Although several studies have applied NAS techniques to GNN design \cite{ijcai2020p195, gao2022graphnas++, wei2023neural}, most focus solely on optimizing task accuracy while neglecting on-device efficiency.
Both factors are crucial for real-world applications, especially in resource-constrained edge scenarios.
In this paper, we propose HGNAS, an efficient hardware-aware graph neural architecture search framework for edge devices to handle real-time inference requirements.
Specifically, we focus on two critical efficiency metrics for edge applications: inference latency and peak memory usage.

\begin{figure}[t]
    \centering
    \includegraphics[width = 0.8\linewidth]{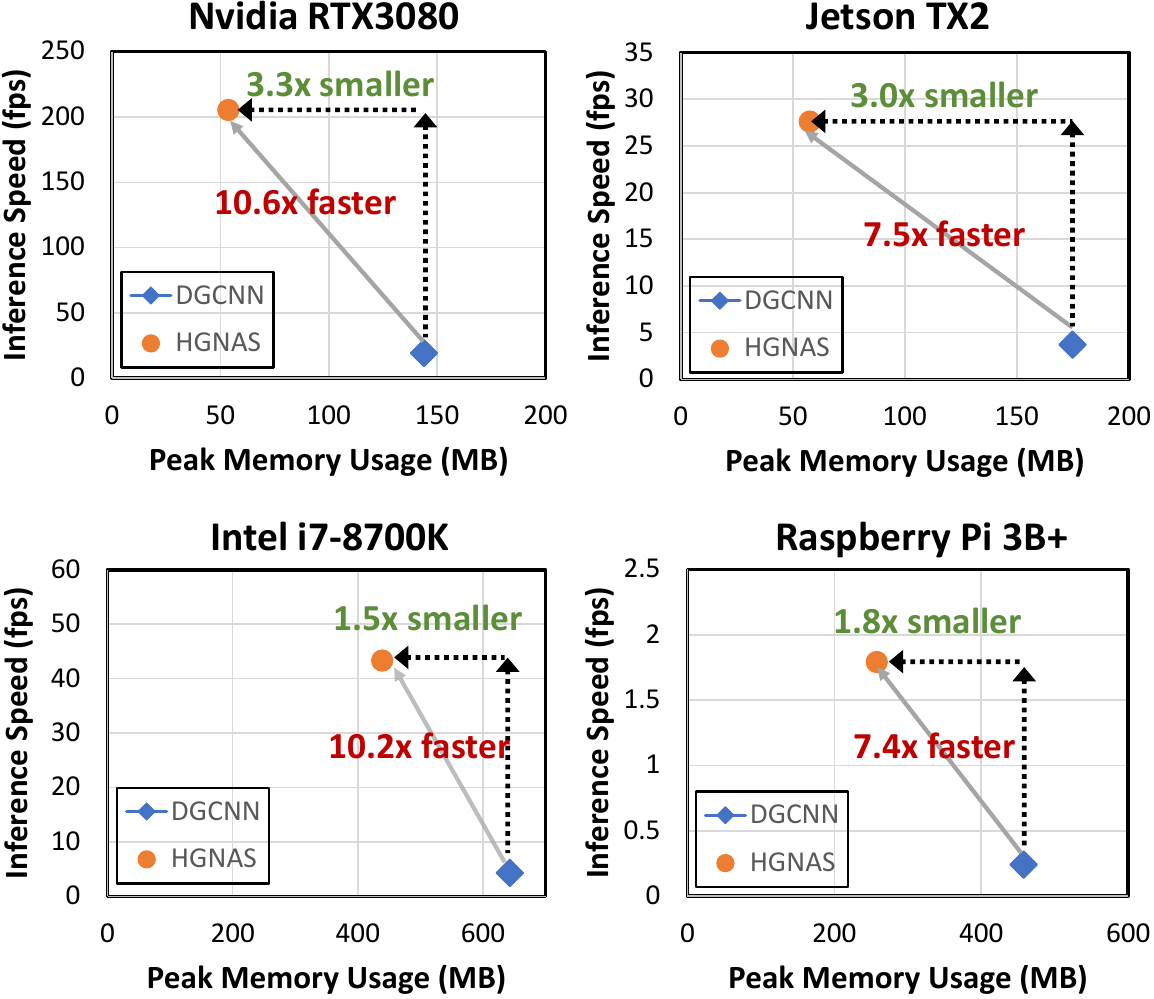}
    \caption{Inference speed vs. peak memory usage. Our approach significantly improves hardware performance while maintaining similar accuracy to DGCNN on ModelNet40. Details are highlighted in bold in Table~\ref{tab:total}.}
    \label{fig:intro-comparison}
    \vspace{-9pt}
\end{figure}

In practice, efficient and high-quality GNN architecture exploration faces many challenges. 
\textbf{1) Tradeoff between efficiency and effectiveness of hardware-aware approaches}.
The hardware efficiency of GNNs is influenced by various factors, including model structure, hardware sensitivity, and graph characteristics~\cite{yan2020characterizing}.
As such, hardware performance awareness strategies that estimate performance using approximate metrics, such as FLOPs, are often inaccurate~\cite{luo2022surgenas}.
While real-time on-device measurement offers more accurate results, the high overhead (communication time, on-device inference time, etc.) in evaluating numerous sub-architectures can severely impede exploration efficiency. 
\textbf{2) Redundancy in layer-wise GNN design space.}
The design manner of stacking the same GNN layer results in operational redundancy, negatively impacting on-device inference efficiency \cite{li2021towards}. 
\textbf{3) Poor search efficiency.}
NAS is often criticized for its lengthy search times, particularly due to the lack of efficient exploration methods for fine-grained GNN search spaces tailored for edge applications.

To address the above issues, HGNAS integrates a novel hardware-aware technique to enable efficient GNN performance prediction.
By abstracting the GNN architecture into graphs, the hardware-aware problems for GNNs can be transformed into graph-related problems, which GNNs are specialized in handling.
This gives rise to the innovative concept of \textbf{Using GNN to perceive GNNs}.
By leveraging a well-polished GNN-based predictor, HGNAS can effectively perceive the latency and peak memory usage of GNN candidates. 
In addition, we provide an efficient peak memory estimation method, leveraging GNN inference profiling to further promote the scalability and robustness of HGNAS.
Furthermore, we develop a fine-grained design space composed of fundamental operations to unleash the potential of GNN computations. 
To improve search efficiency, HGNAS utilizes an efficient multi-stage hierarchical search strategy, mitigating the complexity inherent in exploring the fine-grained design space.
As shown in Fig.~\ref{fig:intro-comparison}, HGNAS has been proven superior in both latency and peak memory usage across various edge devices.
The contributions of this paper can be summarized as follow:
\begin{itemize}
    \item To the best of our knowledge, HGNAS is the first NAS framework to perform efficient graph neural architecture search for resource-constrained edge devices. HGNAS can automatically explore GNN models with multiple objectives (accuracy, latency, and peak memory usage) for targeted platforms.  
    \item We propose an efficient GNN hardware performance predictor that perceives the latency and peak memory usage of GNNs on target devices in milliseconds. To our best understanding, HGNAS is also the first work to achieve hardware performance awareness for GNNs across platforms.
    \item We provide a comprehensive analysis of memory usage during GNN inference and introduce a peak memory estimation method to improve the robustness and scalability of HGNAS.
    \item We propose an efficient multi-stage hierarchical search strategy to accelerate exploration in the expansive GNN fine-grained design space. 
    \item We evaluate HGNAS on four edge devices. As expected, it can achieve up to $10.6\times$ inference speedup and $82.5\%$ peak memory reduction with a negligible accuracy loss on point cloud classification tasks.
\end{itemize}

The rest of the paper is organized as follows. Section \ref{sec:motivation} introduces the background preliminaries and motivations of this paper. Section \ref{sec:methodology} elaborates on the proposed HGNAS framework and Section \ref{sec:experiment} demonstrates the experimental results. 
After that, we discuss the related works in Section~\ref{sec:relatedwork}.
Finally, we conclude this paper in Section \ref{sec:conclusions}.

\begin{figure}[t]
    \centering
    \includegraphics[width = 0.8\linewidth]{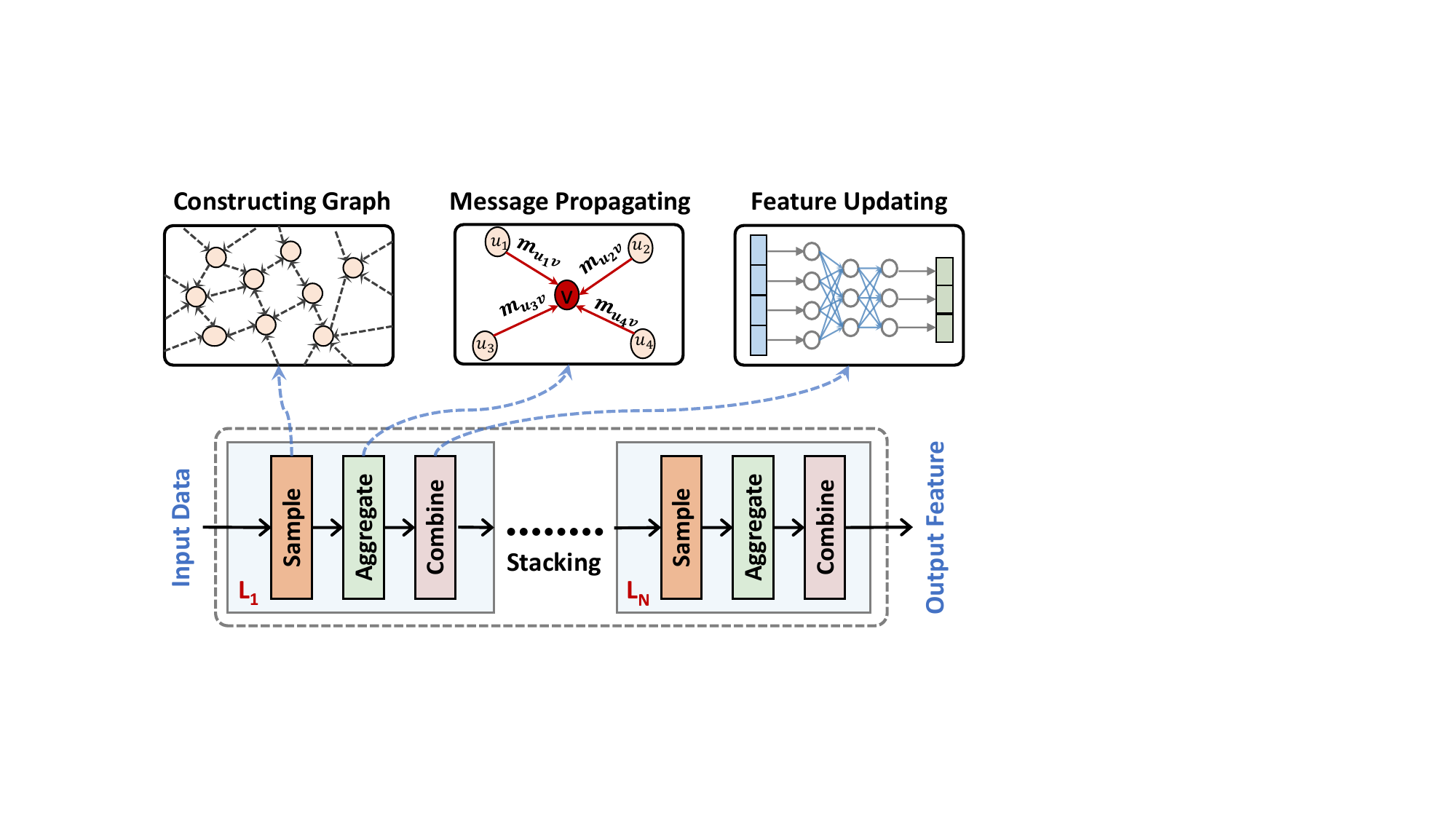}
    \caption{Typical GNN pipeline with MP paradigm.}
    \label{fig:MP}
     \vspace{-9pt}
\end{figure}

\section{Preliminaries and Motivations}\label{sec:motivation}

\subsection{Graph Neural Networks}\label{sec:preliminaries}

Generally, GNN architecture design follows the Message Passing (MP) paradigm. Fig.~\ref{fig:MP} illustrates a typical GNN pipeline using the example of DGCNN \cite{wang2019dynamic}. Each GNN layer consists of \textit{sample}, \textit{aggregate}, and \textit{combine} operations. Specifically, the \textit{sample} operation constructs the graph from the point cloud data for processing, the \textit{aggregate} operation facilitates message propagation among nodes, and the \textit{combine} operation updates all node features. 
Note that \textit{sample} operations are also executed during the inference process to extract the graph structure from the point cloud data.
The propagation rule of DGCNN at layer $k$ is defined as follows:
\begin{equation} \label{equ:edgeconv}
    x_{i}^{k}= \sum _{j \in {{\cal N}{(i)}}}h^{(k)}_{\Theta} \left( x_{i}^{(k-1)} \left\| \left(x_{j}^{(k-1)}-x_{i}^{(k-1)}\right) \right. \right),
\end{equation}
where $h^{(k)}_\Theta$ denotes a neural network, i.e. a MLP.
A GNN model is constructed by stacking GNN layers sequentially.

While this design approach is straightforward, its rigid sequence of operations and repetitive layer stacking limit the scope for innovation, thereby constraining potential performance breakthroughs in GNN models.

\subsection{Observations and Motivations}

\begin{figure}[t]
    \centering
    \includegraphics[width = 0.9\linewidth]{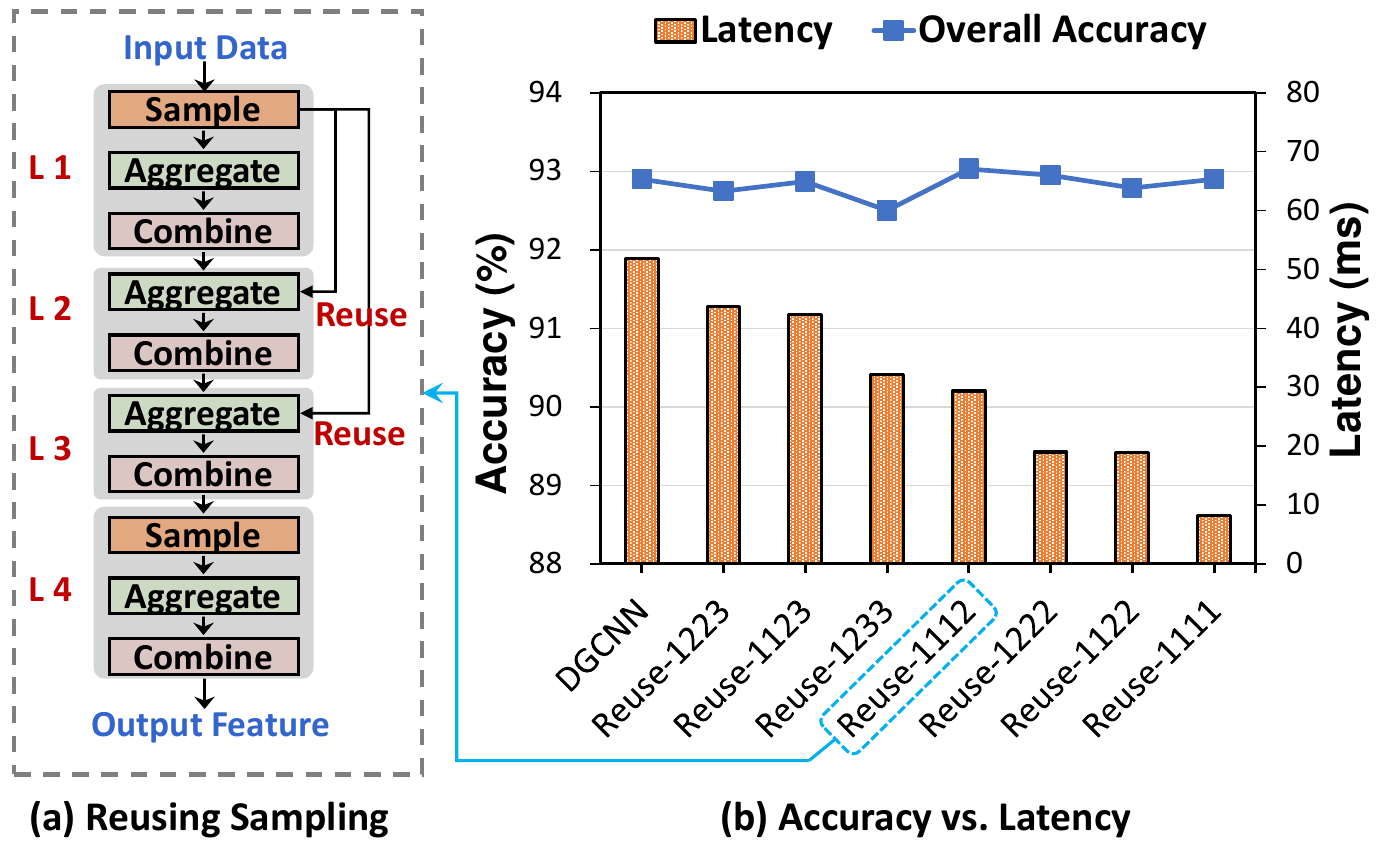}
    \caption{(a) Example of reusing sampling results between layers (Reuse-1112). (b) Accuracy and latency comparison when performing sampled results reuse among different DGCNN layers on ModelNet40~\cite{wu20153d} dataset.}
    \label{fig:ob1}
     \vspace{-6pt}
\end{figure}

Drawing on prior research, we offer key observations that inspire efficient GNN exploration for edge devices.

\textbf{Observation 1: Redundant operations introduced in GNN architecture design bring significant overhead.}
As previously described, GNN models designed following the MP paradigm are formed by stacking GNN layers which possess a fixed sequence of operations.
This naturally raises the question: Do all the elements within each layer contribute to the final performance of the GNN model?
To demonstrate the redundancy, we conduct inter-layer reuse experiments with DGCNN on Nvidia RTX3080. 
Specifically, we remove the \textit{sample} operations within the latter GNN layers and reused the sampling results from the front layers. 
An example of this reuse is shown in Fig.~\ref{fig:ob1}(a).
The experimental results, depicted in Fig.~\ref{fig:ob1}(b), demonstrate that reusing the sampling results has no significant impact on accuracy but remarkably enhances the computational efficiency.
This observation is consistent with the phenomenon found in \cite{li2021towards}, fully demonstrating the considerable overhead introduced by redundant operations in GNN design, posing a major obstacle to GNN computational efficiency optimization.

To eliminate redundant operations in GNNs, \cite{li2021towards, tailor2021towards} propose identifying and manually simplifying the model structure through numerous ablation experiments and analyses, achieving notable speedup.
However, this manual optimization requires extensive trial-and-errors and heavily relies on specialized expertise. 
On the other hand, some researchers decoupled the MP paradigm to construct a more flexible design space, achieving SOTA performance \cite{cai2021rethinking, zhang2022pasca}.
Inspired by the GNN paradigm decoupling, we aim to decouple the GNN layer into operations and construct a fine-grained design space, allowing various configurations (e.g., aggregation range, operation order) to be generated through learning instead of manual effort.

\begin{figure}[t]
    \centering
    \includegraphics[width = 0.9\linewidth]{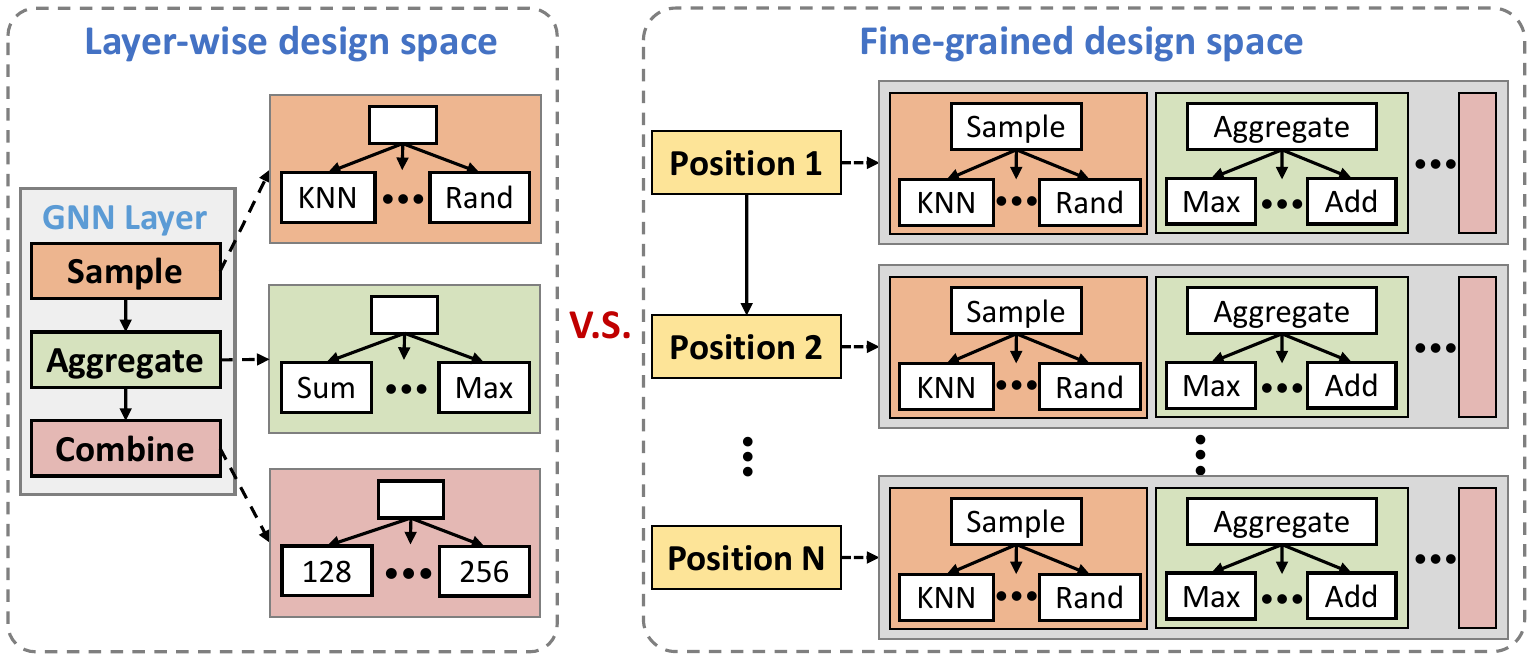}
    \caption{Fine-grained design space greatly expands the scope of architectural exploration.}
    \label{fig:motivation-ob2}
\end{figure}

\textbf{Observation 2: Exploring the GNN fine-grained design space is costly.}
Fig.~\ref{fig:motivation-ob2} illustrates a comparison between the fine-grained design space and the layer-wise design space, highlighting the remarkable expansion of the architectural exploration scope.
Specifically, the fine-grained design space comprises fundamental GNN operations, which covers sub-architectures that grow exponentially with the number of  positions.
For example, the backbone of DGCNN consists of four GNN layers, with each layer comprising three fundamental operations: \textit{sample}, \textit{aggregate}, and \textit{combine}. 
Therefore, to cover most DGCNN variants, the fine-grained design space contains at least $12$ positions, $3$ candidate operations, and at least $N$ functions for each operation (details in Sec.~\ref{sec:design-space}).
Consequently, the design space contains a staggering ${(3N)}^{12}$ possible configurations, thereby exacerbating the complexity of exploration.

The expansion of the GNN design space increases flexibility but also complicates exploration, making manual optimization challenging.
As an AutoML technique, NAS can automatically locate the optimal design without navigating all candidates. 
However, NAS continues to grapple with significant efficiency challenges, exemplified by instances requiring up to 2,000 GPU-days for a single search \cite{zoph2017neural}.
As such, exploring the fine-grained design space requires a more efficient search strategy.
In practice, the contribution of different layers towards the overall accuracy within a model may vary significantly \cite{sheng2022larger, tailor2021towards}. 
Leveraging this insight, we aim to improve exploration efficiency by decoupling the design space and guiding search orientation at different positions.

\begin{figure}[t]
    \centering
    \includegraphics[width = 0.9\linewidth]{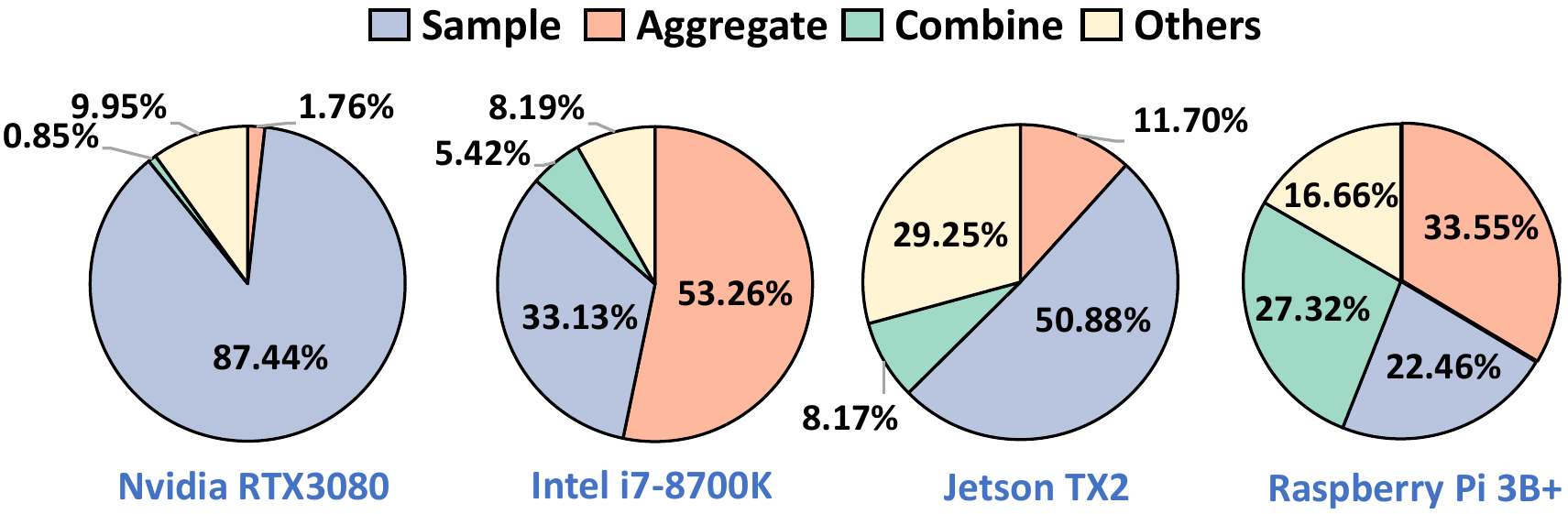}
    \caption{Execution time breakdown of DGCNN across various edge devices on ModelNet40.}
    \label{fig:motivation-ob3}
    \vspace{-9pt}
\end{figure}

\textbf{Observation 3: The same GNN model may behave differently across various computing platforms.}
A detailed breakdown of DGCNN execution time across platforms is illustrated in Fig.~\ref{fig:motivation-ob3}, using data obtained by PyTorch Profiler.
For Nvidia RTX3080 and Jetson TX2, the \textit{sample} operation occupies the majority of execution time. 
This is because GPUs are better at handling compute-intensive matrix operations, and not so good at memory-intensive graph sampling operations.
For Intel i7-8700K, \textit{aggregate} and \textit{sample} dominate the execution time, due to irregular memory accesses patterns.
These observations confirm that DGCNN execution is largely \texttt{I/O-bound} on these platforms.
Conversely, the resource constraints of the Raspberry Pi result in a \texttt{compute-bound} execution, as all three operations are time-consuming.
Therefore, GNN models running on different devices exhibit varying hardware sensitivities that must be carefully considered during architecture design.

In practice, inference efficiency and accuracy are equally important for GNN design in edge scenarios\cite{li2021towards}.
Additionally, the hybrid execution mode of GNNs, which consists of both memory-intensive and compute-intensive operations, poses great challenges for effectively perceiving GNNs hardware performance~\cite{yan2020characterizing}.
Recent studies have explored analytical estimation to improve hardware efficiency for hardware/algorithm co-designs~\cite{zhang2021g, odema2023magnas, zhou2022model, liu2019eslam, li2022eventor, yang2021s}. 
However, these approaches are often tailored for specific hardware architectures, limiting their applicability across diverse edge computing platforms~\cite{benmeziane2021comprehensive}.
Therefore, an efficient and scalable hardware-aware approach is highly desirable.
Inspired by \cite{dudziak2020brp}, a GNN-based hardware performance predictor is integrated to efficiently and accurately perceive GNN hardware performance across various platforms.
\section{Methodology}\label{sec:methodology}

\begin{figure*}[t]
    \centering
    \includegraphics[width = 0.9\linewidth]{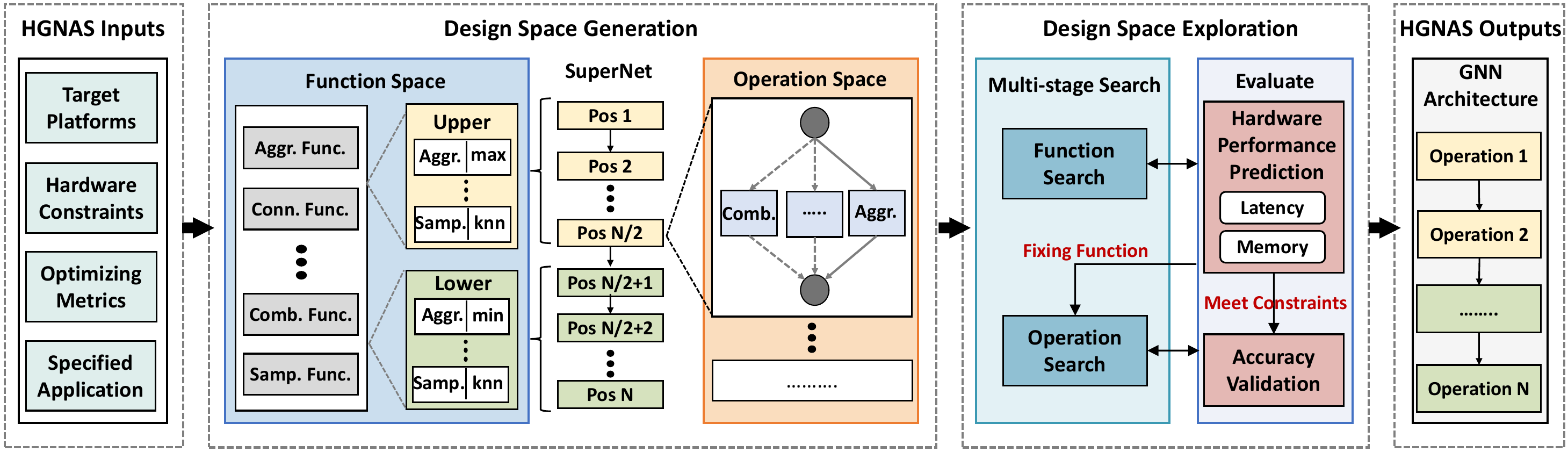}
    \caption{Overview of the proposed HGNAS framework.
    HGNAS aims to search for top-performing GNN architectures that excel in both accuracy and efficiency.}
    \label{fig:framework-overview}
    \vspace{-9pt}
\end{figure*}

\subsection{HGNAS Overview}\label{sec:overview}
This section provides an introduction to the proposed HGNAS framework, as illustrated in Fig.~\ref{fig:framework-overview}.
We begin with the problem definition, in which we integrate hardware performance metrics to guide the exploration process, ensuring that the designed GNNs are both efficient and accurate.
Given the target edge devices, hardware constraints, specified dataset, and the optimizing metrics, HGNAS will first generate a \textit{fine-grained hierarchical design space}, comprising the \textit{Function Space} and \textit{Operation Space}.
HGNAS subsequently constructs a supernet to cover the GNN design space, allowing the search and training processes to be decoupled by one-shot exploration.
Afterward, HGNAS navigates the hierarchical design spaces using the proposed \textit{multi-stage hierarchical search strategy}.
During exploration, each candidate architecture is evaluated based on both its accuracy on the validation dataset and its hardware performance on the target device.
GNN hardware performance is assessed using our proposed \textit{GNN hardware performance predictor}, obviating the need for laborious on-device measurements.
In the following sections, we will detail the key components of the HGNAS framework.

\subsection{Problem Definition}

\begin{table}[t]
\caption{Notations of the involved mathematical symbols and corresponding descriptions.}
\label{tab:Meaning}
\centering
\begin{tabular}{|p{1cm}<{\centering}|p{6cm}|}
\hline
\textbf{Term} & \textbf{Description} \\ \hline
$\cal A$    &     GNN architectures\\ \hline
$\cal{H}$     &     target edge device  \\ \hline
$Lat$     &     inference latency  \\ \hline
$PM$     &     peak memory usage  \\ \hline
$C_{lat}$     &     latency constraint  \\ \hline
$C_{mem}$     &     peak memory constrain  \\ \hline
$acc_{val}$     &     validation accuracy  \\ \hline
$\mathcal{E}$     &     hardware efficiency  \\ \hline
$N$   &     number of positions in GNN supernet \\ \hline
$\mathcal{F}_{\text{obj}}$ & objective function during search \\ \hline
$\alpha$    &     scaling factor for accuracy  \\ \hline
$\beta$    &     scaling factor for efficiency  \\ \hline
\end{tabular}
\end{table}

For simplicity, we first introduce the notations that will facilitate the subsequent exposition, as illustrated in Table~\ref{tab:Meaning}.
In this paper, we aim to co-optimize the accuracy and hardware efficiency of GNNs deployed on edge devices.
Hardware efficiency for GNNs is quantified by two key metrics: inference latency $Lat$ and peak memory usage $PM$, both of which are critical considerations for edge applications.
Given a target edge device $\cal{H}$, along with latency constraint $C_{lat}$, and the peak memory constraint $C_{mem}$, we can formulate the multi-objective optimization task for HGNAS as follows:
\begin{equation}
 \arg \mathop {\max }\limits_{\{{\cal A}, \cal{H}\} } \left( \alpha  * {acc_{val}}\left({{\cal W}^ * }, {\cal A} \right) - \beta  * {\mathcal{E}} \left({\cal A}, \cal{H} \right) \right) ,
    \label{equ1}
\end{equation}
\begin{equation}
    \begin{split}
            s.t. \quad & {{\cal W} ^ * } = \arg \mathop {\max }\limits_{\mathcal W}{acc_{train}} \left( {\cal W}, {\cal A} \right) \\
            & {Lat < C_{lat}} \quad and \quad PM < C_{mem}
    \end{split},
    \label{equ2}
\end{equation}
where $\cal W$ denotes the model weights, $acc_{train}$ is the training accuracy, $acc_{val}$ is the validation accuracy, $\mathcal{E}$ is the hardware efficiency on targeted platform $\cal{H}$, $\cal A$ is the GNN architecture candidate, $\alpha$ and $\beta$ are scaling factors employed to balance the optimization objectives between accuracy and efficiency.
This ensures that the designed GNN aligns with the specific requirements of the target application.
Note that both latency and peak memory usage are jointly influenced by the GNN architecture as well as the platforms where such networks are deployed.
\begin{figure}[t]
    \centering
    \includegraphics[width = 0.8\linewidth]{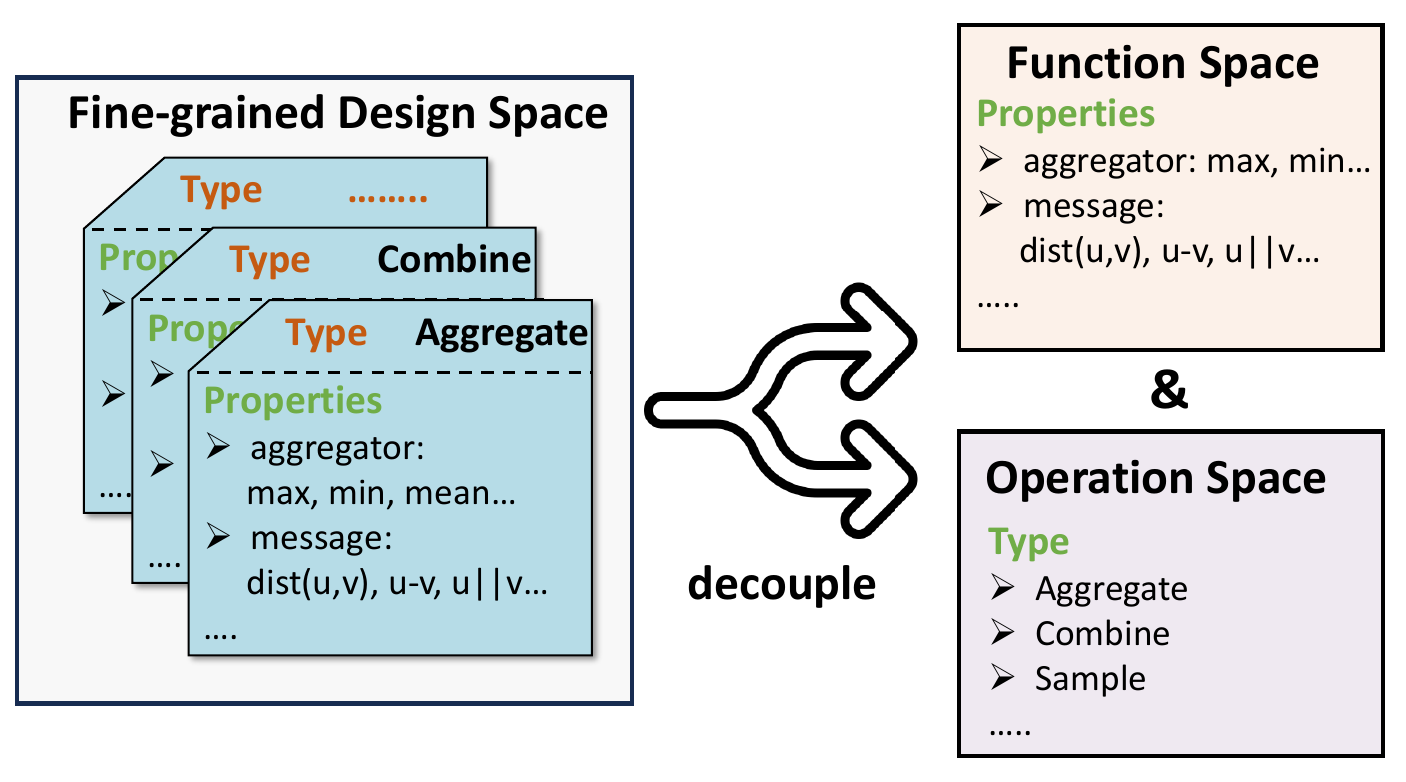}
    \caption{Design space decoupling for GNN.}
    \label{fig:framework-designSpace}
    \vspace{-9pt}
\end{figure}

\subsection{Fine-grained Hierarchical GNN Design Space }\label{sec:design-space}

The traditional layer-wise search space aims to find an optimal block and stack multiple instances of it to construct the optimal architecture, which significantly constrains exploration within the design space. 
Additionally, the computational characteristics of GNNs make the sequence of operations significantly influence efficiency~\cite{yan2020characterizing}.
Notably, the sequence identified in the optimal block may not be universally applicable for optimizing all components of the model.
To unleash the potential of efficient GNN design, we propose a fine-grained hierarchical design space, comprising all the fundamental operations involved in GNN computation.
Furthermore, we decouple the operations by organizing the design space into two hierarchical layers: a \textit{Function Space} and an \textit{Operation Space}. 

\textbf{The GNN supernet.}
To lift the restrictions of the traditional GNN design space, HGNAS adopts a more flexible approach by building the design space based on positions for GNN operations, rather than presetting the number of GNN layers. 
Furthermore, as shown in Fig.~\ref{fig:framework-overview}, HGNAS organizes the GNN design space as a supernet based on positions, in order to minimize exploration overhead through the utilization of the single path one-shot NAS methodology~\cite{guo2020single}.
Candidate architectures are generated by choosing and fixing an operation and its corresponding function at each supernet position.
Specifically, for each position in the supernet, there are four fundamental operations including \textit{connect}, \textit{aggregate}, \textit{combine}, and \textit{sample}, each with distinct properties.
Beyond the operations inherited from the MP paradigm, the \textit{connect} operation, which includes both direct and skip-connections, further enhances the design freedom in constructing GNN models.
Besides, the message type properties in aggregate operations dictate the construction method of messages to be aggregated for point cloud processing.
In practice, supernet training demands uniform lengths for hidden dimensions across all operations situated at each position.
For dimension alignment, HGNAS adds linear transformations to operations, such as \textit{sample} and \textit{aggregate}, that are otherwise incapable of modifying hidden dimensions.
These linear transformations will be omitted in the finalized architecture to avoid introducing additional overhead.

\textbf{The hierarchical design space.} In practice, operations in the fine-grained design space can be further decoupled. 
For example, the \textit{aggregate} operation encompasses various properties, including the aggregation operator and message construction type.
Consequently, HGNAS decouple the fine-grained GNN design space into \textit{Operation Space} and \textit{Function Space}. 
As illustrated in Fig.~\ref{fig:framework-designSpace}, the \textit{Operation Space} includes operation types, while the \textit{Function Space} comprises specific operation properties.
In addition, these two sub-spaces can be explored separately by leveraging the proposed multi-stage hierarchical search strategy (see Sec.~\ref{sec:strategy}) to reduce exploration complexity.
All candidate operations and functions are listed in Table~\ref{tab:space}.

\subsection{Multi-stage Hierarchical Search Strategy}\label{sec:strategy}

\begin{table}[t]
  \centering
  \renewcommand\arraystretch{1.3}
  \caption{The available choices in GNN's supernet.}
  \resizebox{0.9\linewidth}{!}{
    \begin{tabular}{l|l}
    \hline
    \textbf{Operation} & \textbf{Function} \\
    \hline
    Connect & Skip-connect, Identity \\
    \hline
    \multirow{2}[4]{*}{Aggregate} & Aggregator type: sum,  min,  max, mean \\
    \cline{2-2}   & Message type:  Source pos, Target pos,  Relative pos,\\ & Source $\parallel$ Relative pos, Target $\parallel$ Relative pos,\\ & Euclidean distance, Full \\
    \hline
    Combine & $8$, $16$, $32$, $64$, $128$, $256$ \\
    \hline
    Sample & KNN, Random \\
    \hline
    \end{tabular}
  }
  \label{tab:space}
\end{table}

To tackle the excessive exploration complexity introduced by fine-grained design space (see \textbf{Observation 2}), we propose an efficient multi-stage hierarchical search strategy that divides the search process into two stages, corresponding to \textit{Function Space} and \textit{Operation Space}.
Inspired by \cite{guo2020single}, we employ a one-shot approach during the two-stage search process to decouple the supernet training and architecture search, avoiding the exorbitant cost of sub-architecture retraining.
In particular, full supernet training is only performed once after determining the optimal function settings. 
During the search phase, candidates meeting hardware constraints are evaluated for accuracy by inference on the validation set using the pre-trained weights of the supernet.
As illustrated in Alg.~\ref{gnnalgo}, the algorithm inputs include: (1) population size $P$ in evolutionary search, (2) hardware constraints $\cal C$ which will determine the upper bounds of latency and peak memory usage in the finalized GNNs, (3) the target edge device $\mathcal{H}$, (4) \textit{Operation Space} $\mathbb{S}_{op}$, \textit{Function Space} $\mathbb{S}_{f}$, (5) maximum iterations $T$ which determines the terminating conditions, and (6) the number of positions $N$ which affects the exploration scope.
During exploration, HGNAS first searches for an optimal function setting in the \textit{Function Space}.
After that, the GNN supernet is pre-trained based on this optimal setting.
Subsequently, a multi-objective search is conducted across all positions in the supernet for optimal operations.
Finally, the algorithm outputs the top-performing model $\cal A^*$ tailored for the target edge device $\mathcal{H}$.
The details of the multi-stage search strategy are outlined below.

\begin{algorithm}[t]
\textbf{Inputs:} population size $P$, hardware constraints $\cal C$, target device $\mathcal{H}$, \textit{Operation Space} $\mathbb{S}_{op}$, \textit{Function Space} $\mathbb{S}_{f}$, max iteration $T$, number of positions $N$.

\textbf{Outputs:} the best found GNN design $\cal A^*$ for target device $\mathcal{H}$.

Initialize GNN supernet ${\cal N}_{super}$ with $N$ positions and two function sets $upper \leftarrow \mathbb{\O}$, $lower \leftarrow \mathbb{\O}$ \\
{/* Stage 1: Function search */} \\

Assign function set: \parbox[t]{0.5\linewidth}{
${\cal N}_{super}[0, N/2] \leftarrow upper$, \\
${\cal N}_{super}[N/2+1, N] \leftarrow lower$}

\For{$1 \leq t \leq T$}{
$\{upper, lower\} \leftarrow EA(P, {\cal N}_{super}, \mathbb{S}_{f}, obj = max(acc_{val}))$\\
}
Fix function set $\mathbb{F}  \leftarrow \{upper,lower\}$ for ${\cal N}_{super}$\\ 
Re-initialize and pre-train ${\cal N}_{super}(\mathbb{S}_{op}, \mathbb{F})$\\
{/* Stage 2: Operation search */} \\
Initialize operation set $\cal O \leftarrow \mathbb{\O} $\\
\For{$1 \leq t \leq T$}{
    ${\cal O} \leftarrow EA({\cal P}, {\cal N}_{super}, \mathbb{F}, S_{op},
    obj= max({\cal F}_{obj}({\cal C})))$\\
}
\textbf{return} optimal architecture ${\cal A^*} \leftarrow \{{\cal O}, \mathbb{F}\}$

\caption{\small Multi-stage hierarchical search strategy.}
\label{gnnalgo}

\end{algorithm}

\textbf{Stage 1: Function search.} 
In this stage, HGNAS seeks to identify a function setting that maximizes supernet accuracy.
During the search, HGNAS utilizes an evolutionary algorithm ($EA$) to iteratively select sub-functions from the GNN supernet.
The score of each sub-function is determined by the corresponding supernet accuracy, requiring only a few training epochs.
To further improve the exploration efficiency, HGNAS partitions the $N$ positions of the GNN supernet into two halves, sharing one set of functions among the \textit{Upper} half ($0,...,N/2$) and another set among the \textit{Lower} half ($N/2+1,...,N$). 
This sharing scheme is inspired by the differing contributions of the front and latter GNN layers to accuracy~\cite{tailor2021towards}. 
Although this approach risks overlooking some promising architectures, the considerable gain in exploration efficiency justifies its use.
For a supernet with $12$ positions, through sharing functions among positions in the decoupled design space, HGNAS can reduce the number of exploration candidates from $\bm{4.2 \times {10^{12}}}$ to $\bm{1.7 \times {10^7}}$.
Finally, an optimal function set $\mathbb{F}$ is determined for initializing the supernet ${\cal N}_{super}$. 
Note that fixing $\mathbb{F}$ will significantly reduce the complexity of subsequent exploration.

\textbf{Stage 2: Operation search.}
Upon fixing $\mathbb{F}$, we pre-train the GNN supernet to obtain shared weights for all sub-architectures, thus avoiding retraining.
During this stage, HGNAS explores the remaining \textit{Operation Space} with the aim of locating a set of operations to maximize the accuracy and efficiency of the GNN candidate on the target device.
Specifically, the objective function during the operation search is formulated as:
\begin{equation}
    \mathcal{F}_{\text{obj}}(\mathcal{C}) = \left\{
    \begin{aligned}
        &0, & \text{if } \mathcal{E} \ge \mathcal{C} \\
        &\alpha * acc_{val} - \beta * \mathcal{E}, & \text{if } \mathcal{E} < \mathcal{C}
    \end{aligned}
    \right. 
    \label{equ3}
\end{equation}
where $\mathcal{E}$ represents the hardware efficiency, including latency and peak memory usage.
For GNN architectures that fail to meet the hardware constraints $\cal C$, we will not further evaluate the accuracy and directly mark zero to avoid the unqualified GNNs.
The evaluation of hardware efficiency is based on the proposed GNN hardware performance predictor (see Sec.~\ref{sec:predictor}), which can perceive the latency and peak memory usage of candidate GNNs on the target device in milliseconds.
By adjusting $\alpha$ and $\beta$, we can easily direct the search trend (towards more accurate or more efficient) to serve the requirements of different application scenarios.

\begin{figure}[t]
    \centering
    \includegraphics[width = 0.95\linewidth]{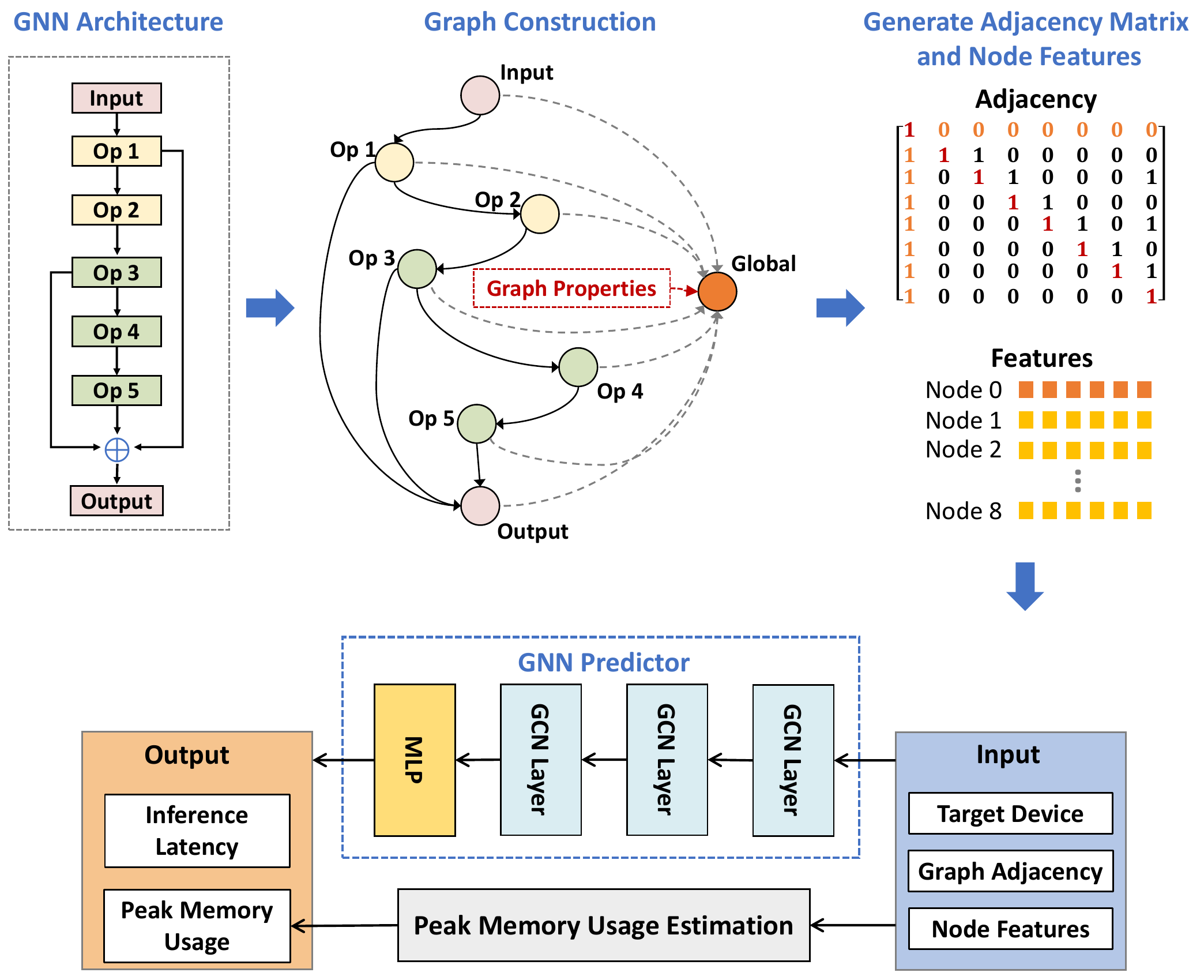}
    \caption{Latency and peak memory usage prediction of a candidate GNN model for the target device.}
    \label{fig:predictor}
    \vspace{-9pt}
\end{figure}

\subsection{GNN Hardware Performance Prediction}\label{sec:predictor}

By abstracting the GNN architecture into graphs, the GNN hardware-awareness problem can be reformulated as a graph representation learning problem, an area where GNNs excel.
As such, we propose an efficient GNN hardware performance predictor to learn the relationship between GNN architectures and hardware efficiency.
In addition, we perform an intensive analysis of the GNN computation process and propose a peak memory usage estimation method to assist in predicting GNN's run-time peak memory usage.
As shown in Fig.~\ref{fig:predictor}, the prediction process consists of the following phases: \textit{graph construction}, \textit{node feature generation}, \textit{latency prediction}, \textit{peak memory usage prediction}.
Note that the proposed GNN hardware performance predictor is only applied during the search process, whereas the experiment results in Sec.~\ref{sec:experiment} are directly measured on target edge device for fair comparisons.

\textbf{Graph construction.} 
During this phase, HGNAS abstracts GNN architectures into directed graphs, which serve as the input for the GNN predictor. 
While GNNs typically use undirected graphs for better connectivity, we choose directed graphs because their unique dataflow properties significantly impact prediction accuracy.
Specifically, the nodes in these architecture graphs represent inputs, outputs, and operations, while the edges depict the dataflow within the GNN architecture.
In practice, accurate prediction of hardware efficiency requires both candidate model architecture and the graph property of the input dataset, which GNN execution highly depends on.
However, this plain abstraction of the original GNN architectures is too sparse for the predictor to obtain sufficient structural features, and lacks the necessary information on input data.
To address this limitation, HGNAS introduces a global node that connects with all the other nodes in the graph to improve the graph connectivity.
In this way, the propagation of operational information throughout the entire architecture graph is significantly improved, which benefits the learning of the GNN predictor.
Finally, HGNAS will output an architecture graph in adjacency matrix format $G \in \mathbb{R}^{N \times N}$, where $N$ denotes the number of nodes.

\textbf{Node feature generation.} 
For the node feature initialization, HGNAS employs the one-hot strategy commonly used in GNNs \cite{kipf2017semi} to encode the possible candidates at each position.
For an operation node, the node feature comprises the operation type and its corresponding function.
Specifically, HGNAS encodes these two components into $7$-dimensional and $9$-dimensional one-hot vectors, respectively, and subsequently concatenates these vectors to form the node feature.
For input and output nodes, HGNAS assigns them with zero vectors, indicating that they do not have any specific operation associated with them.
Regarding the global node, HGNAS encodes the input graph data properties (such as the number of nodes, density, etc.) into a $16$-dimensional vector as the global node feature.
Afterwards, a node feature matrix $X \in \mathbb{R}^{N \times L}$ will be generated as input to the GNN predictor, where $L$ is the feature length.

\textbf{Latency prediction.}
To avoid the over-smoothing problem often induced by deeper GNNs on small-scale graphs (i.e., the abstracted architecture graph), the latency predictor consists of only three GCN layers~\cite{kipf2017semi} and a multi-layer perceptron (MLP).
Specifically, the GCN layers utilize the sum aggregator with hidden dimensions of $256 \times 512 \times 512$.
For the MLP part, we employ three fully-connected layers with hidden dimensions of $256 \times 128 \times 1$ respectively, followed by a LeakyReLU activation function $\sigma$ for generating a scalar prediction of latency.
The predictor takes target edge device $\mathcal{H}$, adjacency matrix $\mathcal{G}$, and node feature matrix $\mathcal{X}$ as inputs, and outputs the predicted latency.
A one-hot encoding method is used for the target device input.  
When integrating new devices, only a small amount of data collection is required, followed by incremental training of the predictor.
The GNN-based latency prediction can be formulated as follows:
\begin{equation}
Lat \left( \mathcal{A} \right) = \sigma \left(MLP\left(GCN\left(\mathcal{A},\mathcal{G},\mathcal{X},\mathcal{H}\right)\right)\right) 
\end{equation}

The predictor is trained for $250$ epochs on $30$K randomly sampled candidate architectures ($21$K for training and $9$K for validation) in our fine-grained GNN design space, with labels obtained from measurement results on various edge devices.
During the predictor training, we utilize the \textit{mean absolute percentage error} (MAPE) as the loss function to mitigate the impact of potential outliers, such as system disruptions during data collection.
Additionally, we utilize \texttt{AdamW} as the training optimizer, employing a batch size of $32$ and an initial learning rate of $0.0008$, which is dynamically adjusted using the \texttt{ReduceLROnPlateau} scheduler.
As architecture graphs typically contain only a few dozen nodes, the prediction overhead is mostly negligible, requiring only about 10 ms on GPU platforms.

\textbf{Peak memory usage prediction.}
In the design of efficient GNNs for edge applications, peak memory usage serves as a critical metric, substantially influencing the feasibility of such networks on the target platform.
Therefore, we extend the GNN hardware performance predictor to include peak memory usage prediction during the search process, thereby mitigating the risk of unsuitable design choices.
For data preparation, we collect $30$K random GNN architectures from our fine-grained GNN design space, employing the same dataset splitting strategy for latency prediction.
For GPU devices, we utilize the \texttt{max\_memory\_allocated} function in PyTorch \cite{memory} to obtain peak memory usage data.
For CPU devices, we perform data collection using the {\texttt{Resource}} tool \cite{resource}.
As we observed, data collection accuracy directly affects prediction accuracy.
Hence, we independently sample the peak memory data of each architecture to minimize interference.
Furthermore, we employ the same predictor structure and training scheme as for latency prediction, while adjusting the initial learning rate to $0.0003$ and batch size to $16$.
In practice, initializing the peak memory predictor using the weights of the latency predictor can expedite the model convergence, thereby suggesting an interrelation between the two performance metrics (see Sec.~\ref{sec:cor-relationships}).
As such, we utilize the well-trained latency predictor's weights to initialize the peak memory predictor.
This approach enables the peak memory predictor to reach training completion within tens of epochs while attaining performance on par with the latency predictor.

\textbf{Peak memory usage estimation.}
Unlike the runtime latency of GNNs, influenced by multiple factors, peak memory usage is closely associated with the size of the tensor generated during inference.
This correlation allows us to develop an accurate method for estimating peak memory usage for a wider range of platforms with enhanced evaluation robustness.
To this end, we profile several randomly sampled architectures from the fine-grained design space and summarize the rules governing the patterns of their peak memory usage.
During GNN inference, memory usage can be broadly divided into three categories: those related to the model, the dataset, and intermediate variables.
The model-related memory usage $M_p$ is approximately equal to $U_k \times N_p$, where $U_k$ and $N_p$ are the precision and number of the model parameters.
And dataset-related Memory usage $M_d$ is determined by the volume of data required for the inference of a single batch.
Upon loading both the model and the data, the total memory usage $M$ is given by $M = M_p + M_d$.
In addition, the memory usage of intermediate variables are generated during each operation in the forward execution.
For clarity, we henceforth use $M$ to denote post-execution memory usage and $M'$ for pre-execution memory usage of the current operation.
Specifically, the memory consumption brought by some typical GNN operations are delineated below.

(1) \textbf{\textit{Sample.}} For this operation, the intermediate result takes the form of a graph that is represented as an edge list.
Assume that there are $N$ nodes in the graph and $K$ neighbors are sampled for each node, the memory usage of \textit{sample} operation can be formulated as follows:
\begin{equation}
M_{sample} = N_e \times 2 \times U_{index},
\end{equation}
where $N_e$, which equals to $N \times K$, is the number of edges in the graph and $U_{index}$ is the precision of the edge index. 
After executing \textit{sample} operation, the memory usage $M$ is equal to $M' + M_{sample}$.

(2) \textbf{\textit{Aggregate.}}The \textit{aggregate} operation is executed in two phases: message construction and message broadcasting. The former generates messages on the edges, while the latter updates the node features through message passing. 
By assuming the node feature length as $L$, the introduced memory usage can be calculated as follows:
\begin{equation}
M_{msg} = N_e \times 2 \times L \times U_{k},
\end{equation}
\begin{equation}
M_{broad} = N \times L \times U_{k}.
\end{equation}
Upon completing the message construction phase, the updated memory usage $M$ is equal to $M' + M_{msg}$.
Additionally, the memory usage after the completion of the message construction phase is denoted as $M_{mc}$ for subsequent peak memory calculations.
In practice, memory allocated for messages is automatically recycled after message broadcasting. 
Consequently, the memory usage $M$ is updated after the \textit{aggregate} operation as follows: $M = M' + M_{broad}- M_{msg}$.

(3) \textbf{\textit{Combine.}} This operation is typically implemented based on MLP, introducing memory usage that can be straightforwardly calculated as follows:
\begin{equation}
M_{com} = N \times L_{out} \times U_{k},
\end{equation}
where $L_{out}$ is the output feature dimension of MLP. 
Upon execution of this operation, the updated memory usage $M$ becomes $M' + M_{com}$.

The total memory usage of a candidate GNN architecture can be determined by accumulating the memory consumed in each successive operation during forward computation.
Nevertheless, the peak memory usage during GNN inference may not necessarily align with $M$.
Explicit message construction in graphs with numerous edges and high-dimensional node features can consume considerable memory, not accounted for in $M$. 
In practice, this phase is often where peak memory usage is observed.
Thus, GNN's final peak memory usage can be calculated as follows:
\begin{equation}
    PM = Max(M, M_{mc}^{(1)}, ..., M_{mc}^{(n)}),
\end{equation}
where $n$ is the number of \textit{aggregate} operations in GNNs.

In summary, this estimation method allows for direct calculation of peak memory during GNN inference across different GPU devices.
Note that this approach does not directly assess peak memory on CPU devices, as memory usage on these devices encompasses both program execution data and Tensor data.
Nevertheless, the method remains useful for assessing the relative peak memory consumption of candidates during exploration, irrespective of the device targeted.
Practically, HGNAS integrates both the predictor output and estimation metrics to improve evaluation robustness throughout the search process.
When the predictor output is lower than the estimated value, indicating a biased prediction, the estimated value is used to measure the peak memory usage of the candidate architecture.
\begin{figure*}[t]
    \centering
    \includegraphics[width = 1\linewidth]{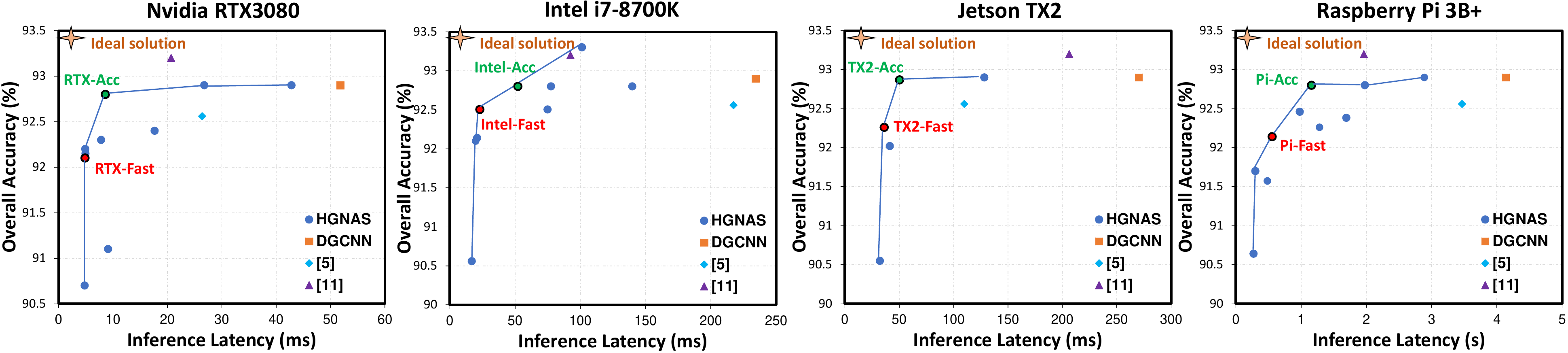}
    \caption{Comparison between existing networks and HGNAS across various devices.}
    \label{fig:pareto}
    \vspace{-9pt}
\end{figure*}

\section{Experiment}\label{sec:experiment}
\subsection{Experimental Settings} \label{sec:setting}

\textbf{Baselines and datasets.}
To evaluate HGNAS, we consider two different application datasets for the graph classification task: the point cloud processing benchmark ModelNet40~\cite{wu20153d} and the text analysis dataset MR~\cite{ZhangYCWWW20}.
Evaluation and hyper-parameter settings are based on~\cite{tailor2021towards} and~\cite{wei2023neural}, respectively. 
Our comparison included several baselines: (1) the popular point cloud processing model DGCNN~\cite{wang2019dynamic}, (2) two manually optimized variants of DGCNN~\cite{li2021towards, tailor2021towards}, (3) two SOTA methods for optimizing point cloud classification accuracy PointGS~\cite{jiang2023pointgs} and LGFNet~\cite{wu2024context}, and (4) the GNN NAS framework PAS \cite{wei2023neural}.
For a fairer comparison with manual optimizations in~\cite{li2021towards,tailor2021towards}, we adopt their reported accuracy, inference speedup, and memory reduction as the baseline on the GPU platform.
For other edge platforms, we reproduce these baselines based on PyTorch Geometric (PyG) framework ~\cite{Fey/Lenssen/2019}, due to the lack of pre-trained models and evaluation results.
To compare with PAS, we use the PAS\_G model, which is the relatively lightweight optimal searched GNNs, for our testing.
In addition, all experimental and profiling results are obtained using the PyG framework, taking the average results of $10$ runs.

\textbf{HGNAS settings.}
We assign $12$ positions for the GNN supernet to cover DGCNN architectures. 
For design space exploration, the maximum number of iterations is set to $1000$, and the population size for the evolutionary algorithm ($EA$) is fixed at $20$. 
Both the search and training phases of HGNAS are carried out on an Nvidia V100 GPU.
During function search and operation search, the number of GNN supernet training epochs is set as $50$ and $500$, respectively.

\textbf{Predictor settings.} 
The predictor is trained for $250$ epochs using $30$K randomly sampled architectures from our fine-grained design space, where $21$K of these are adopted for training and $9$K for validation.
Labels for these architectures are obtained from performance measurements on various edge devices.
The performance of the predictor is evaluated using MAPE and validation accuracy as metrics.

\textbf{Edge devices.}
We employ four edge devices for comparing HGNAS and competitors: (1) Nvidia RTX3080 with $10$GB memory, (2) Intel i7-8700K, (3) Jetson TX2 with $8$GB memory, (4) Raspberry Pi 3B+ with a Cortex-A5 processor and $1$GB memory. 
Note that the hardware performance of both competitors and HGNAS is derived from real GNN inference executions on the specified devices.
\subsection{Evaluation on ModelNet40}

\begin{figure}[t]
    \centering
    \includegraphics[width = 0.9\linewidth]{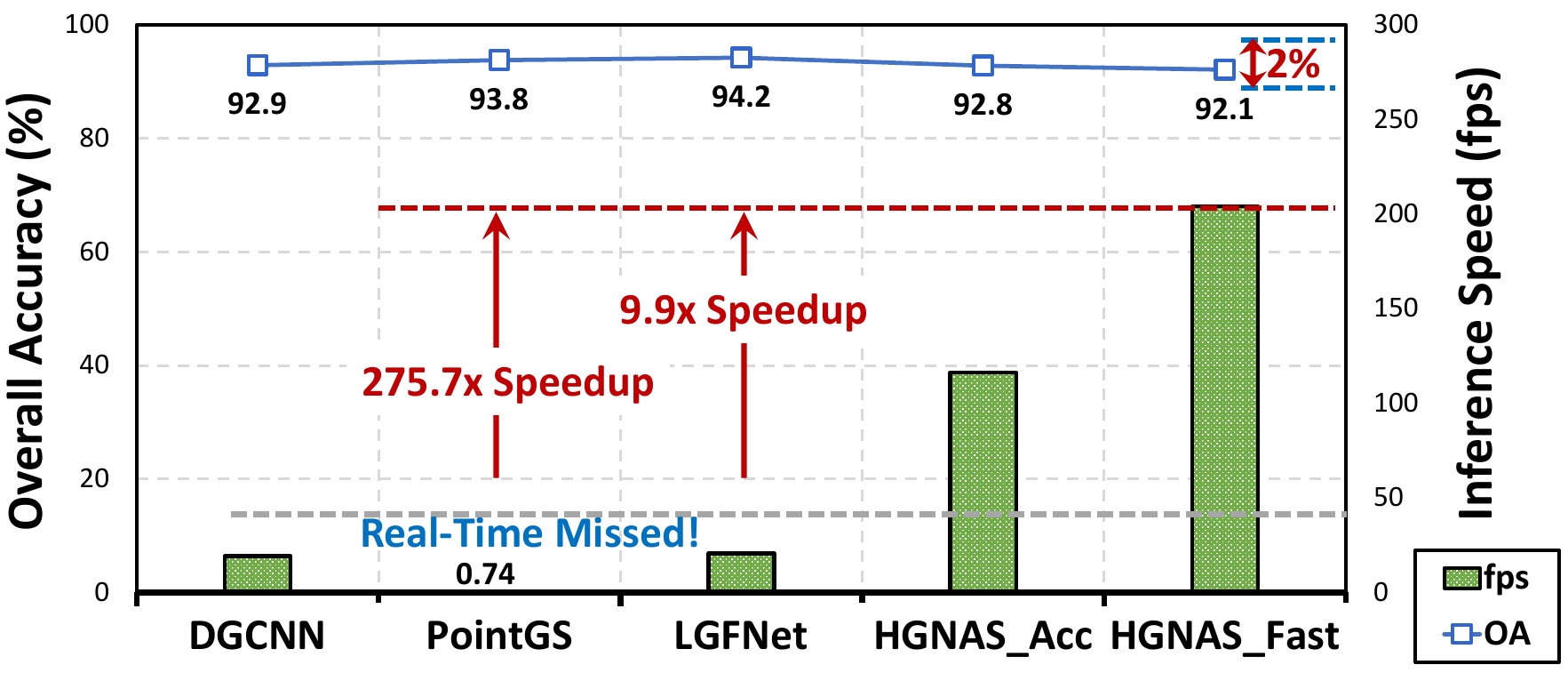}
    \caption{Accuracy and efficiency trade-offs in GNN designs, highlighting substantial gains with minimal accuracy loss.
    }
    \label{fig:sota_compare}
    \vspace{-9pt}
\end{figure}

\subsubsection{Accuracy vs. Efficiency} \label{sec:accvseff}
Fig.~\ref{fig:pareto} depicts the outcome of HGNAS exploration, aiming for reduced latency and enhanced accuracy.
The ideal solution is designated by a star and positioned in the top-left corner of the figure.
The green markers labeled Device\_Acc (e.g., RTX\_Acc) represent architectures optimized for specific devices by HGNAS without compromising accuracy.
In contrast, the red markers labeled Device\_Fast permit a $1\%$ drop in accuracy.
Note that the original intention of HGNAS is to design efficient GNN models for edge devices. 
As such, we have applied a larger scaling factor $\beta$ to the hardware efficiency metrics during search, with the objective of identifying ultra-efficient architectures that still adhering to accuracy requirements.
The results demonstrate that HGNAS consistently maintains a better performance frontier on various devices, which is guaranteed by the accurate hardware performance prediction of the candidate GNNs during the search.
By carefully selecting scaling factors, HGNAS can easily balance hardware efficiency and task accuracy, as detailed in Sec.~\ref{sec:tradeoff}.

In real-time applications, efficiency is as crucial as accuracy. 
Significant progress in enhancing GNN accuracy does not mitigate their primary deployment obstacle on edge devices: inefficient inference.
As shown in Fig.~\ref{fig:sota_compare}, PointGS and LGFNet prioritize optimizing accuracy, significantly improving point cloud classification. 
However, this emphasis on accuracy compromises inference efficiency, leading to higher on-device latency and failing to meet real-time requirements.
Conversely, HGNAS incorporates efficiency metrics into the NAS objective, enabling the creation of a GNN model that achieves inference efficiency gains with minimal accuracy loss, meeting real-time constraints.
Specifically, HGNAS demonstrates a balance between accuracy and efficiency, with less than 2\% accuracy loss compared to PointGS and LGFNet, while achieving speedups of 275.7$\times$ and 9.9$\times$ on the RTX3080 platform, respectively.

\begin{table}[t]
  \centering
  \setlength{\tabcolsep}{2pt}
  \renewcommand\arraystretch{1.5}
  \caption{Comparison of HGNAS and existing methods, where OA and mAcc denote overall accuracy and balanced accuracy, respectively.}
  \resizebox{\linewidth}{!}{
    \begin{tabular}{c|c|c|c|c|c|c}
    \hline
    Device & Network &Size [MB] & OA   & mAcc & Latency [ms] & PM [MB]   \\
    \hline
    \multirow{5}[2]{*}{RTX} & DGCNN & 1.81 & 92.9 & 88.9 & 51.8 & 174.9   \\
         & \cite{li2021towards}& - &   92.6   &  89.6   &  (2.0$\times$$\uparrow$)   &    (51.9\%$\downarrow$)   \\
         & \cite{tailor2021towards}& -  &  93.2    &    90.6  & (2.5$\times$$\uparrow$)     &    -  \\
         & RTX\_Acc & 1.61 & 92.8 & 90.1 & 8.6 (6.0$\times$$\uparrow$) &   60.5 (65.4\%$\downarrow$)   \\
         & \textbf{RTX-Fast} &  \textbf{1.46} & \textbf{92.1} & \textbf{88.5} & \textbf{4.9 (10.6$\times$$\uparrow$)}  &   \textbf{53.8 (69.2\%$\downarrow$)}  \\
         & RTX\_Small & 1.49 & 92.4 & 89.3 & 4.9 (10.6$\times$$\uparrow$) &   30.7 (82.5\%$\downarrow$)   \\
    \hline
    \multirow{5}[2]{*}{Intel} & DGCNN &  1.81 & 92.9 & 88.9 & 234.2 & 643.0      \\
         & \cite{li2021towards} & 2.33&   92.6   &  89.6   &  217.4 (1.1$\times$$\uparrow$)   &   581.3 (9.6\%$\downarrow$)   \\
         & \cite{tailor2021towards} & 1.80 &   93.2   &    90.6  &  92.4 (2.5$\times$$\uparrow$)   &   454.6 (29.3\%$\downarrow$)   \\
         & Intel\_Acc  & 1.61 &   92.8   &  90.1   &   52.2 (4.5$\times$$\uparrow$)  & 426.8(33.6\%$\downarrow$)     \\
         & \textbf{Intel\_Fast} & \textbf{1.47} &   \textbf{92.5}   &   \textbf{88.8}   &    \textbf{23.1 (10.2$\times$$\uparrow$)}  &  \textbf{439.2 (31.6\%$\downarrow$)}    \\
         & Intel\_Small  & 1.37 &   92.2   &  89.6   &   41.6 (5.6$\times$$\uparrow$)  & 398.8(38.0\%$\downarrow$)     \\
    \hline
    \multirow{5}[2]{*}{TX2} & DGCNN & 1.81  & 92.9 & 88.9 & 270.4 &   174.9    \\
         & \cite{li2021towards} & 2.33 &   92.6   &  89.6   &  109.9 (2.5$\times$$\uparrow$)   &   121.2 (30.7\%$\downarrow$)   \\
         & \cite{tailor2021towards} & 1.81 &  93.2    &    90.6  & 206.4 (1.3$\times$$\uparrow$)    & 34.5 (80.3\%$\downarrow$)      \\
         & TX2\_Acc & 1.60 & 92.9 & 89.7 & 50.5 (5.3$\times$$\uparrow$) &   60.5 (65.4\%$\downarrow$)    \\
         & \textbf{TX2\_Fast} &  \textbf{1.48} & \textbf{92.2} & \textbf{88.7} & \textbf{36.3 (7.5$\times$$\uparrow$)} &   \textbf{57.6 (67.1\%$\downarrow$)}   \\
         & TX2\_Small & 1.34 & 92.5 & 89.0 & 88.4 (3.1$\times$$\uparrow$) &   32.6 (81.4\%$\downarrow$)    \\
    \hline
    \multirow{5}[2]{*}{Pi}& DGCNN & 1.81  & 92.9 & 88.9 & 4139.1 &   457.8    \\
         & \cite{li2021towards} & 2.33 &   92.6   &  89.6   &  3466.0 (1.2$\times$$\uparrow$)   &   354.3 (22.6\%$\downarrow$)   \\
         & \cite{tailor2021towards} & 1.81 &  93.2    &    90.6  & 1961.7 (2.1$\times$$\uparrow$)      &   271.1 (40.8\%$\downarrow$)    \\
         & Pi\_Acc & 1.47 &   92.8   &   89.3   &  1165.3 (3.6$\times$$\uparrow$)    &   270.2 (41.0\%$\downarrow$)   \\
         & \textbf{Pi\_Fast} & \textbf{1.36} &    \textbf{92.1}  &    \textbf{88.3} &   \textbf{557.6 (7.4$\times$$\uparrow$)}  &  \textbf{257.8 (43.7\%$\downarrow$)}  \\
         & Pi\_Small & 1.40 &   92.1   &   88.7   &  683.4 (6.1$\times$$\uparrow$)    &   257.8 (43.7\%$\downarrow$)   \\
    \hline
    \end{tabular}
  \label{tab:total}
  }
  \vspace{-9pt}
\end{table}

\subsubsection{HGNAS over Existing Graph Neural Architectures}\label{sec:mn_compare}

In this set of experiments, we benchmark HGNAS against competitors across four edge devices.
For HGNAS, we conduct separate explorations prioritizing accuracy, latency, and peak memory usage by assigning them larger scaling factors.
The results are summarized in Table~\ref{tab:total}. 
It can be observed that GNN models designed by HGNAS show improved hardware efficiency in terms of reduced latency and peak memory usage across various edge computing platforms while maintaining similar accuracy levels.
Specifically, HGNAS achieves speedups of $6.0\times$, $4.5\times$, $5.3\times$, and $3.6\times$, all while maintaining the same accuracy as DGCNN on the four edge devices.
With a permissible $1\%$ accuracy loss, the Device\_Fast GNNs designed by HGNAS achieve speedups of up to $10.6\times$, $10.2\times$, $7.5\times$, and $7.4\times$.
This superior performance is attributed to the accurate latency predictions during the model search process.
In practice, only achieving latency awareness is insufficient to meet edge application requirements, and peak memory awareness is important for memory-constrained scenarios.
Therefore, based on our previously published work~\cite{zhou2023hardware}, we propose two new approaches in this paper: one for peak memory prediction and another for peak memory estimation, to guide the exploration.
By incorporating peak memory metrics into the multi-objective optimization process, the Device\_Small architecture is explored to achieve lower peak memory usage on each device.
As shown in Table~\ref{tab:total}, the Device\_Small architectures achieve peak memory reductions of $82.5\%$, $38.0\%$, $81.4\%$, $43.7\%$ across the four edge devices, outperforms other competitors and the Device\_Fast architectures.
This substantially enhances the viability of deploying GNNs on edge devices with limited resources.
Furthermore, HGNAS realizes a more pronounced reduction in peak memory usage on GPU platforms, like RTX3080 and TX2, primarily due to enhanced evaluation stability gained from the combination of predictor outputs and estimation results.
Compared to the latency-aware design of Device\_Fast, Device\_Small reduces peak memory usage by up to $43\%$, highlighting the effectiveness of the proposed peak memory awareness approach.
Meanwhile, as shown in Table~\ref{tab:total}, GNNs designed by HGNAS aiming for lower latency also exhibit reduced peak memory usage, and vice versa.
This exposes some correlation between the two metrics, which we will analyze in detail in Sec.~\ref{sec:cor-relationships}.
Moreover, despite the numerous candidates introduced by the fine-grained design space, HGNAS achieves search efficiency comparable to other hardware-aware GNN search frameworks by leveraging an efficient multi-stage search strategy.
Specifically, G-CoS~\cite{zhang2021g}, with $10^9$ candidate architectures, requires $4$ GPU hours to complete a search. 
In contrast, HGNAS, with $10^{12}$ candidate architectures, requires only $3$ GPU hours.

\subsection{Evaluation on MR} \label{sec:mr}
\begin{figure*}[t]
    \centering
    \includegraphics[width = 0.9\linewidth]{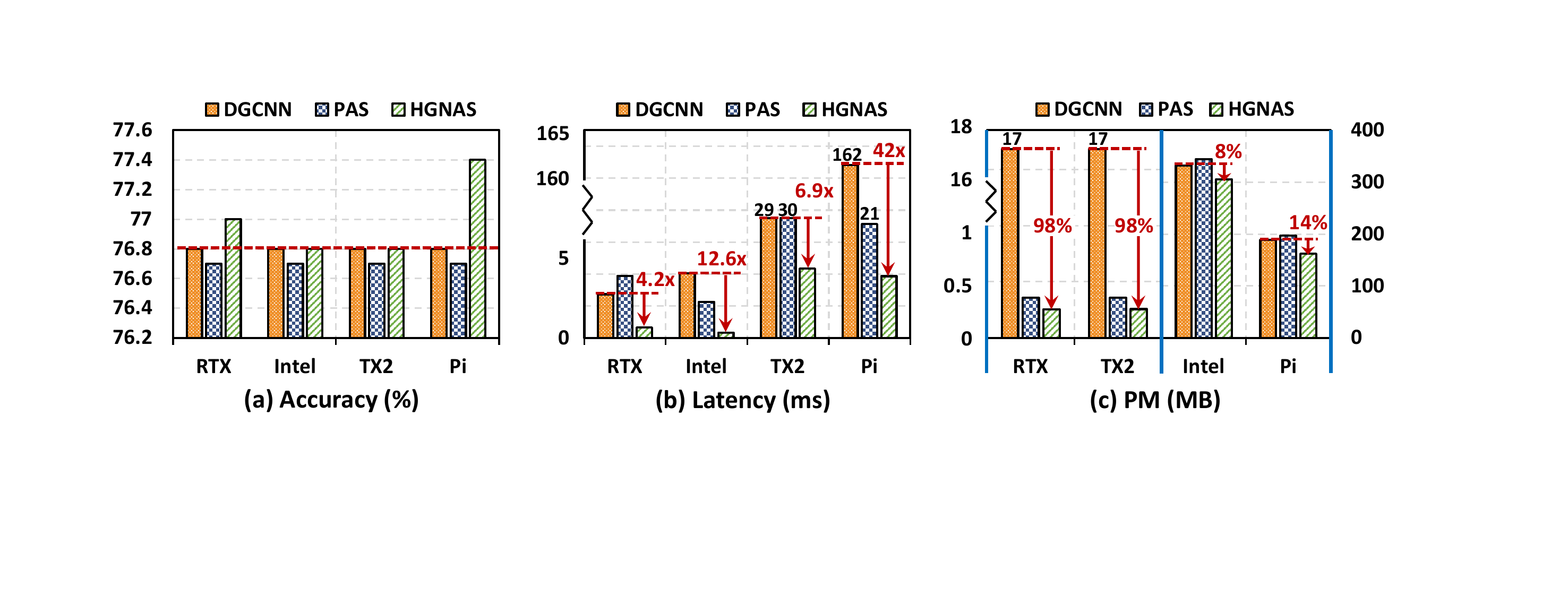}
    \caption{Comparison of HGNAS and other methods on the MR dataset.
    }
    \label{fig:mr}
    \vspace{-9pt}
\end{figure*}

To demonstrate the scalability of HGNAS, we conducted additional experiments using the text analysis dataset MR.
This dataset is also used for graph classification tasks, featuring unique data characteristics, varying numbers of nodes, and different hardware sensitivities compared to the point cloud dataset ModelNet40.
Fig.~\ref{fig:mr} presents the experimental results, with some data scaled to improve readability due to their large variance.
Compared to DGCNN and PAS, our model achieves similar or superior task accuracy, along with improvements in inference efficiency by factors of $4.2\times$, $12.6\times$, $6.9\times$, and $42\times$ on four edge platforms, respectively.
Additionally, we achieved a $98\%$ reduction in peak memory usage on both the RTX3080 and Jetson TX2 platforms, which are equipped with GPU resources.
Moreover, given the resource limitations of the Raspberry Pi, we relaxed hardware constraints during the search process. This resulted in a substantial accuracy improvement, demonstrating HGNAS's ability to balance accuracy and efficiency.
The substantial performance improvement demonstrates HGNAS's advantages in designing efficient GNN models.

\begin{figure}[t]
    \centering
    \includegraphics[width = 0.9\linewidth]{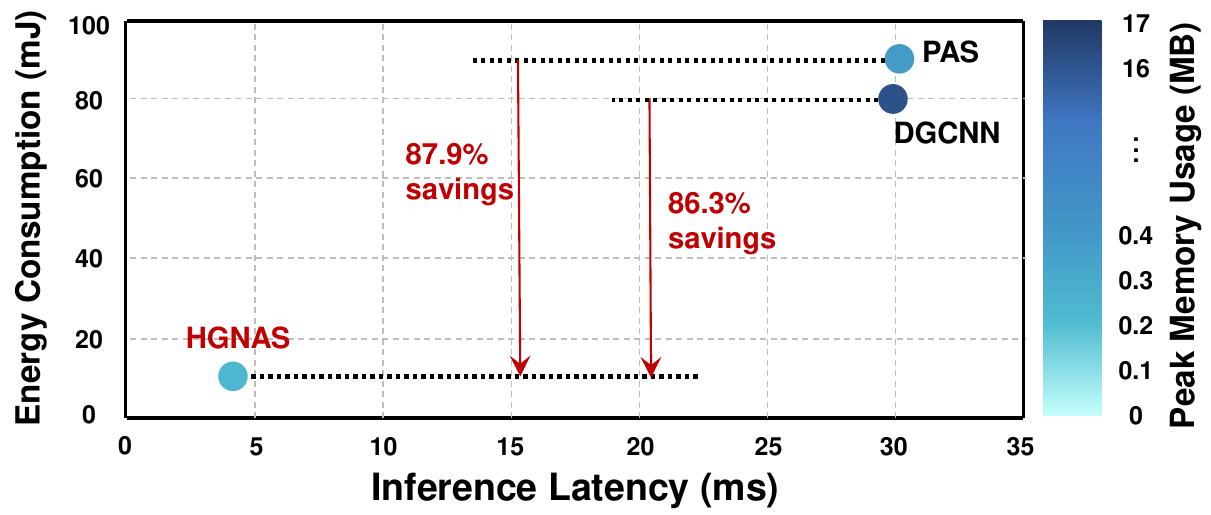}
    \caption{Evaluation of HGNAS and baselines for inference energy consumption on Jetson TX2.
    }
    \label{fig:energy}
    \vspace{-9pt}
\end{figure}

As a pioneering solution for deploying GNNs on edge devices, HGNAS excels in various tasks and easily scales to optimize diverse hardware performance metrics.
In addressing the crucial energy consumption metrics for edge deployments, HGNAS can scale effectively via incremental predictor training.
By integrating energy metrics into the objective function, HGNAS can automatically design energy-efficient GNN models, tailored to various application scenarios.
We conducted an energy-efficient GNN architecture search experiment using the Jetson TX2 platform, and the results are shown in Fig.~\ref{fig:energy}.
Specifically, energy consumption was estimated by measuring the average power during model inference with the Jtop tool and multiplying it by the inference time. 
Compared to DGCNN, which consumes $79.9$ mJ per inference, the GNN designed by HGNAS offers substantial energy savings of $86.3\%$, requiring only $10.93$ mJ per inference, while maintaining similar accuracy.
Compared to PAS, which consumes $90.2$ mJ, HGNAS achieves an $87.9\%$ reduction in energy consumption along with a $6.9\times$ speedup.
In practice, we have observed that models with lower latency typically exhibit lower energy consumption, as also mentioned in~\cite{luo2022surgenas}.
By analyzing 1000 GNN candidates, we established a strong correlation between energy consumption and latency, evidenced by a correlation coefficient of $0.76$.
Based on the principle that $energy = power \times latency$ and given minimal power fluctuations for the same inference task, energy consumption is highly dependent on latency.
Consequently, our latency-centered search process results in optimal GNNs with correspondingly lower energy consumption.

\begin{figure}[t]
    \centering
    \includegraphics[width = 0.8\linewidth]{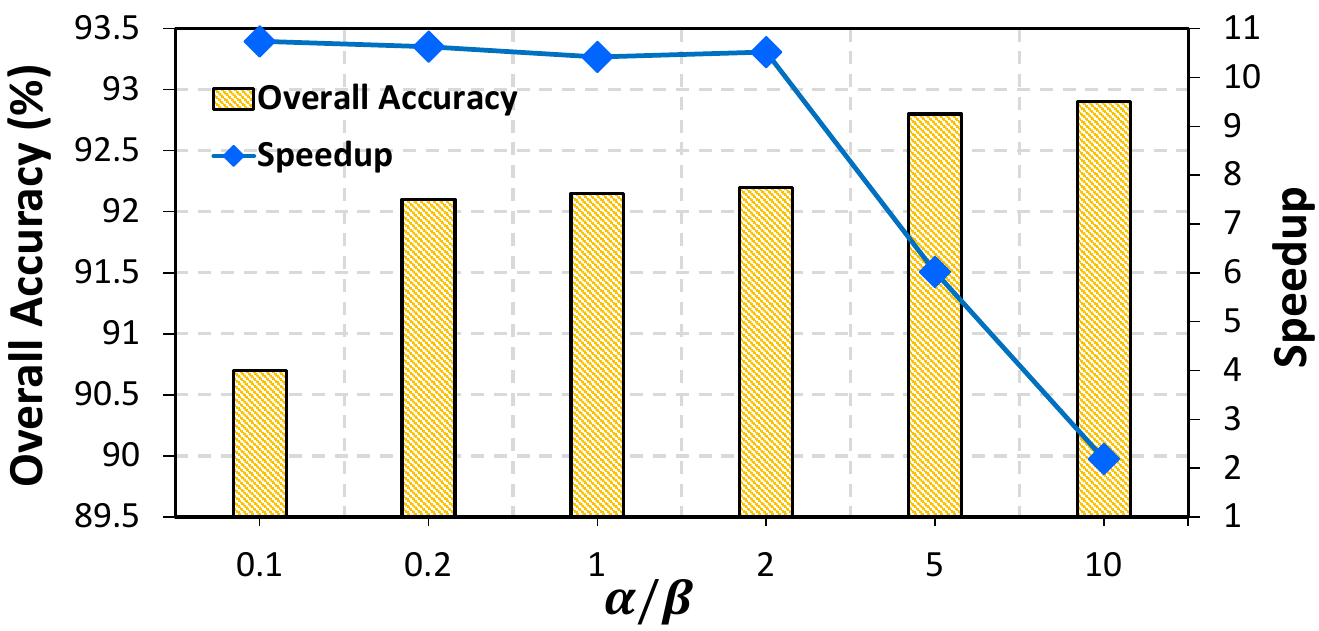}
    \caption{The trade-off between accuracy and efficiency by scaling factor $\alpha$ and $\beta$.
    Efficiency is represented by the speedup when comparing to DGCNN.
    }
    \label{fig:tradeoff}
    \vspace{-9pt}
\end{figure}

\subsection{Trade-off via scaling factors}\label{sec:tradeoff}

In this set of experiments, we perform multiple explorations with various scaling factors on ModelNet40 using an Nvidia RTX3080 to demonstrate the effectiveness of the proposed hardware-aware GNN NAS method.
The results are presented in Fig.~\ref{fig:tradeoff}, where we use the ratio of $\alpha$ and $\beta$ to represent the different search orientations.
Specifically, when $\alpha / \beta$ is smaller, the search results are more in favor of lower latency than higher accuracy.
Conversely, when $\alpha / \beta$ is larger, the search results tend to emphasize more on accuracy.
This indicates that the scaling factors $\alpha$ and $\beta$ can adeptly balance accuracy and latency, demonstrating the efficacy of the suggested hardware-aware architecture search.
In practical applications, this approach offers significant flexibility, as we can dynamically steer the search process towards either accuracy or efficiency by adjusting $\alpha$ and $\beta$ according to the demands of specific tasks.

\begin{figure}[t]
    \centering
    \includegraphics[width = 0.8\linewidth]{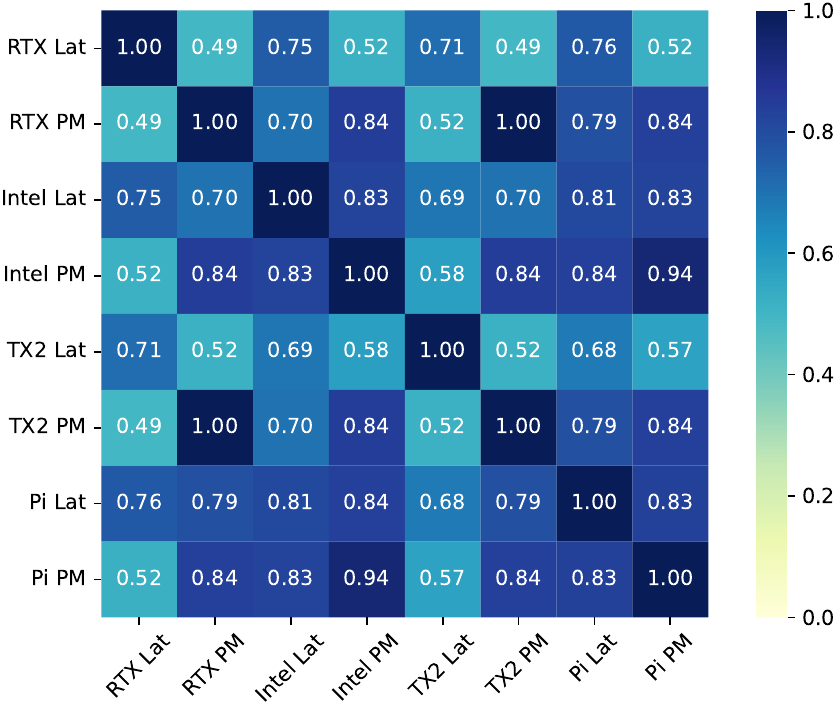}
    \caption{Illustration of the correlation relationships between the latency and the peak memory usage.}
    \label{fig:heatmap}
    \vspace{-9pt}
\end{figure}

\subsection{Correlation between the measured latency and peak memory usage}\label{sec:cor-relationships}

As mentioned before, in HGNAS, we aim to search for resource-efficient GNNs with minimal inference latency and peak memory usage on specific target platforms.
In practice, we observe a strong correlation between these two metrics, in which architectures with lower latency also have smaller peak memory usage.
To illustrate this point, we randomly sample 1000 GNN architectures in the fine-grained GNN design space and subsequently measure their latency and peak memory usage on four edge devices. 
As depicted in Figure~\ref{fig:heatmap}, the correlation coefficient between the measured latency and peak memory usage reaches 0.83, suggesting a strong correlation.
This observation suggests that GNNs exhibiting low latency also tend to be memory-efficient,  especially on edge devices (i.e. Raspberry Pi) with limited computational and memory resources.
As such, we initialise the peak memory predictor using the weights from the latency predictor, leading to a substantial reduction in training time.
Furthermore, a consistent correlation between these metrics is observed across various devices.
This suggests that GNN models optimized by HGNAS for one device are unlikely to exhibit significantly sub-optimal performance on others.
This observation offers a valuable trade-off: designers can aim for sub-optimal performance based on predictions from similar devices without incurring the overheads of seeking extreme optimization for each new platform.

\subsection{Prediction Results}

\begin{figure*}[t]
    \centering
    \includegraphics[width = 0.85\linewidth]{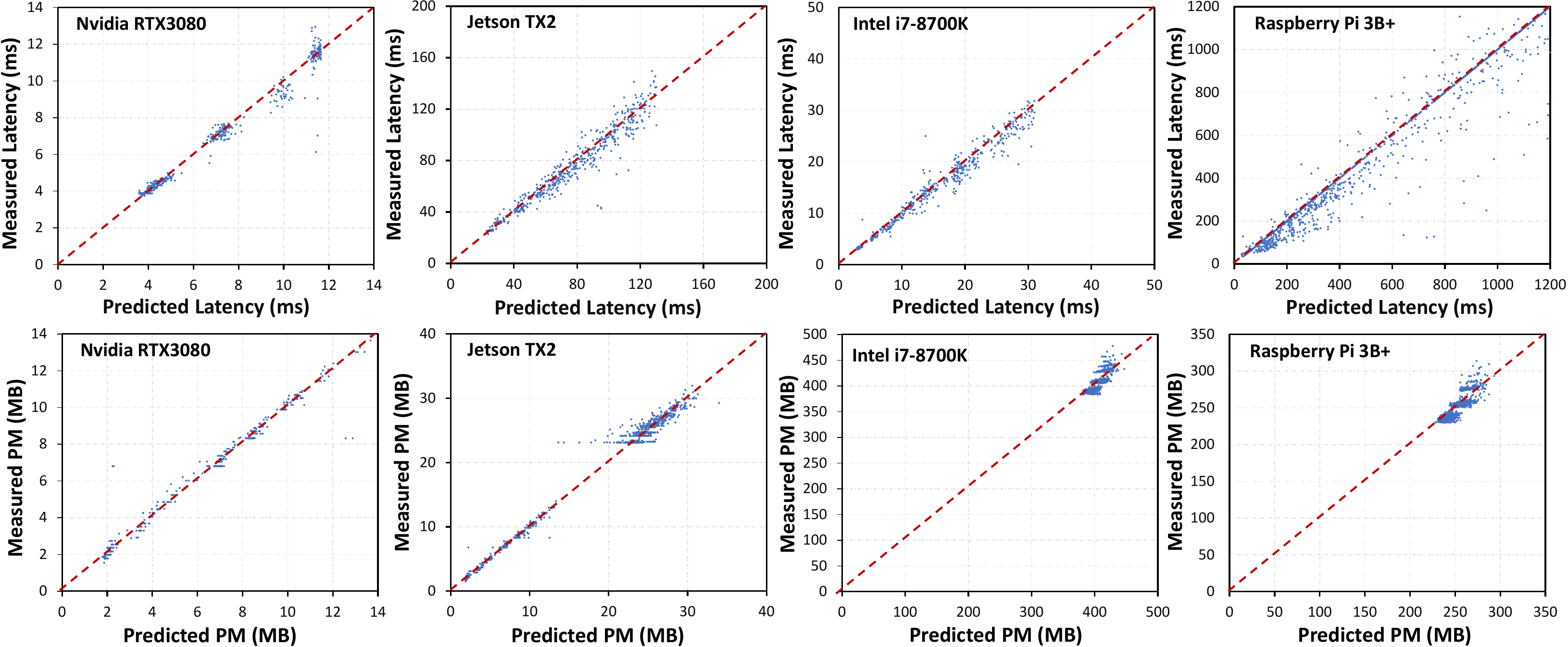}
    \caption{Predictor's accuracy evaluation on various edge devices.
    The measured latency and PM denote the actual results collected from real devices.
    The distance between each blue point and the red dash line reflects the predictor's accuracy.
    }
    \label{fig:sactter}
    \vspace{-9pt}
\end{figure*}

In this set of experiments, we evaluate the effectiveness of the proposed GNN hardware performance prediction approach. 
As illustrated in Fig.~\ref{fig:sactter}, our predictor exhibits high accuracy in predicting hardware efficiency for GNNs across various devices on ModelNet40 dataset.
Specifically, the MAPE for latency predictions is approximately $6\%$ on the RTX3080, Intel i7-8700K, and Jetson TX2. 
However, it rises to around $19\%$ on the Raspberry Pi due to latency measurement fluctuations.
Meanwhile, the MAPE for peak memory predictions is approximately $4\%$ on the RTX3080 and Jetson TX2, compared to about $2\%$ on the remaining devices.
Table~\ref{tab:error_bound} details the prediction results on ModelNet40, showing the percentage of predictions within the stipulated error bounds compared to on-device measurements.
Across diverse devices, the latency prediction method achieves an approximate $80\%$ accuracy with a $10\%$ error bound, while the peak memory usage prediction method consistently surpasses $90\%$ accuracy.
Meanwhile, the peak memory estimation approach achieves a comparable accuracy of more than $90\%$ with a $10\%$ error bound on GPU devices.
Furthermore, the integration of predictor outputs and estimation results fortifies the evaluation robustness, as evidenced by the substantial decrease in peak memory usage for the RTX\_Small and TX2\_Small.
In practice, the GNN predictor performs better for GNNs that have faster inference speeds, thereby aiding HGNAS in searching more efficient GNN designs.
For hardware performance prediction on the MR dataset, HGNAS achieves higher accuracy. The predictor attains an average accuracy of $88.25\%$ for latency prediction with a $5\%$ error bound across four devices. Additionally, the prediction accuracy for peak memory usage is approximately $99\%$.

\begin{table}[t]
\centering
\caption{Performance of the proposed predictor.}
\label{tab:error_bound}
\setlength{\tabcolsep}{6pt}
\renewcommand\arraystretch{1.5}
\resizebox{0.9\linewidth}{!}{
\begin{tabular}{cccccccccc}
\hline
\multirow{2}{*}{\begin{tabular}[c]{@{}c@{}}Error\\ Bound\end{tabular}} & \multicolumn{4}{c}{Latency Accuracy {[}\%{]}} &  & \multicolumn{4}{c}{PM Accuracy {[}\%{]}} \\ \cline{2-5} \cline{7-10} 
& RTX   & Intel    & TX2     & Pi      &  & RTX   & Intel   & TX2    & Pi     \\ \hline
±1\%   & 10.3        & 17.2        & 11.1    & 61.4    &  & 28.5      & 37.8       & 23.7   & 32.9   \\ 
±5\%   & 49.6        & 66.4        & 48.9    & 69.2    &  & 89.1      & 96.0       & 80.2   & 90.1   \\ 
±10\%  & 79.3        & 87.2        & 79.5    & 77.6    &  & 95.8      & 99.9       & 93.8   & 99.6   \\ \hline
\end{tabular}
}
\end{table}

\begin{figure}[t]
    \centering
    \includegraphics[width = 0.95\linewidth]{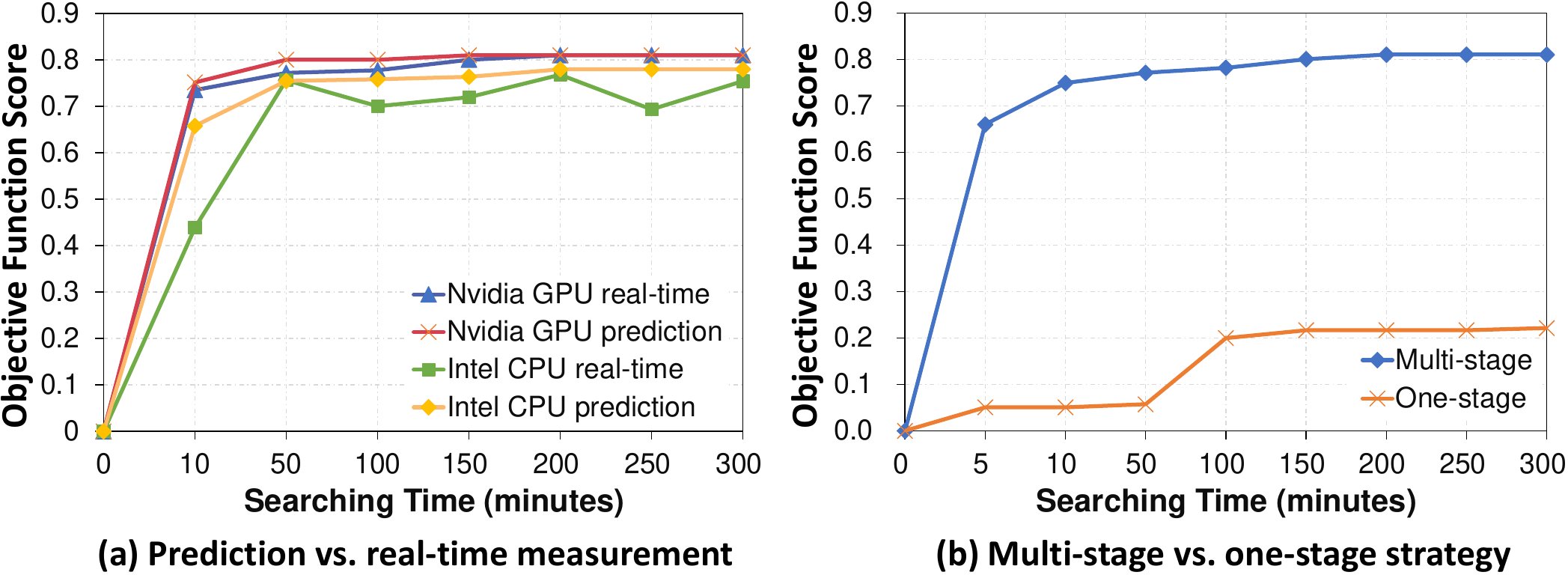}
    \caption{(a) Performance comparison between real-time measured and prediction-based search. (b) Search time reduction with the multi-stage strategy.}
    \label{fig:ablation}
    \vspace{-9pt}
\end{figure}

\subsection{Ablation Studies}

In this set of experiments, we evaluate both the proposed GNN hardware performance prediction method and the multi-stage search strategy for their efficacy in optimizing the exploration of GNN architectures within a fine-grained design space.

\textbf{Prediction vs. real-time measurement.} 
Excessive deployment and communication overheads make real-time measurements of candidate GNNs impractical on most edge devices, whereas our prediction approach assesses them in milliseconds without compromising results.
Fig.~\ref{fig:ablation}(a) illustrates that our GNN prediction approach effectively enhances search efficiency, as the models searched using both methods yield comparable performance.
In particular, our prediction method becomes indispensable when real-time measurements are unfeasible, such as on Raspberry Pi.

\textbf{Multi-stage vs. one-stage search strategy.}
In practice, the expansive fine-grained design space presents formidable challenges for efficient exploration; traditional single-stage search strategies often become mired in a myriad of candidate solutions.
On the other hand, Fig.~\ref{fig:ablation}(b) shows that our multi-stage hierarchical search strategy significantly speeds up the exploration, capable of identifying an optimal GNN architecture within a matter of GPU hours.
Specifically, by leveraging the multi-stage search strategy, HGNAS completes a single exploration in approximately $3$ hours.

\subsection{Insight from GNNs Designed by HGNAS} \label{sec:insight}

\begin{figure}[t]
    \centering
    \includegraphics[width = 0.85\linewidth]{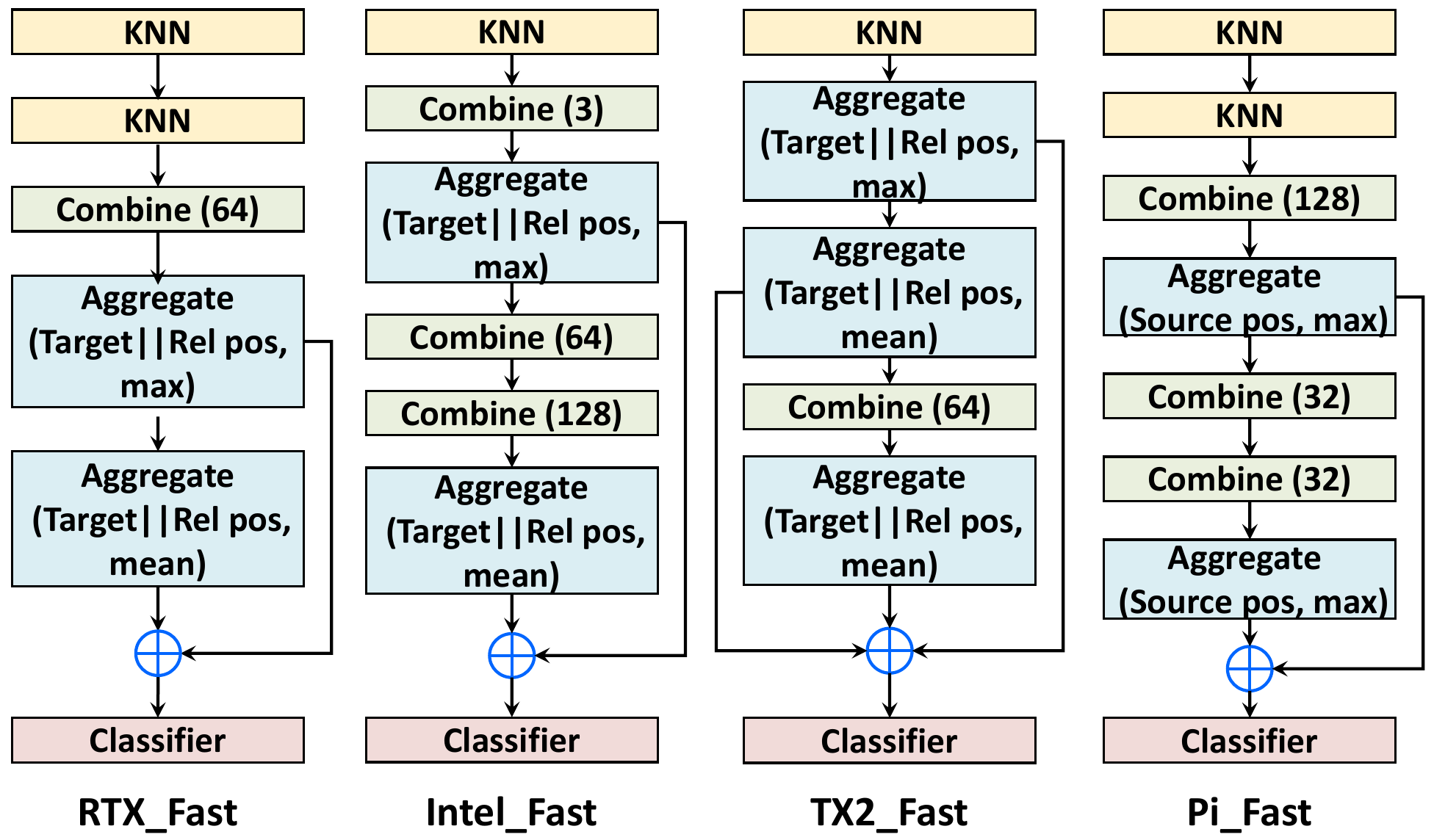}
    \caption{Visualization of GNN models designed by HGNAS.}
    \label{fig:GNNdesigned}
    \vspace{-9pt}
\end{figure}

As demonstrated before, the same GNN model may perform differently across diverse edge computing platforms, which implies that the optimization needed is also device-specific.
By leveraging the proposed GNN hardware performance predictor, HGNAS effectively identifies device-specific characteristics to successfully determine the optimal GNN architectures.
Fig.~\ref{fig:GNNdesigned} visualizes the GNN architectures designed by HGNAS for ModelNet40 dataset.
Note that adjacent KNN operations are merged during execution to avoid invalid graph construction.
The results clearly show that the hardware-efficient architectures designed by HGNAS closely align with the characteristics of the target device, corroborating the GNN computational patterns highlighted in \textbf{Observation 3}.
For example, given that KNN constitutes a significant portion of execution time on RTX3080 and Jetson TX2, GNN models tailored for these platforms incorporate fewer valid KNN operations.
Moreover, the optimal GNNs for Intel CPU involves fewer aggregate operations, while those tailored for the Raspberry Pi prioritize simplifying each operations.

\section{Related Works}\label{sec:relatedwork}

With the success of GNNs in various edge applications~\cite{ying2018graph, qi2017pointnet++, wei2023neural}, there is growing attention in the research community on enhancing inference efficiency in environments with limited resources~\cite{tailor2021towards, shao2021branchy, odema2023magnas}.
\cite{tailor2021towards} performed a thorough analysis of the computational process of GNNs, finding that the initial layers contribute most significantly to task accuracy, with diminishing returns observed in later layers. 
They proposed enhancing inference efficiency by integrating a stronger feature extractor in the early stages and simplifying the later stages. 
Additionally, \cite{li2021towards} noted that sampling constitutes the major overhead in GNNs, with frequent duplication of results in later layers from the initial layer. 
They suggested that reusing sampling results from the first layer could significantly accelerate inference. 
Nevertheless, manually designing such efficient GNNs requires substantial trial-and-errors, inevitably resulting in significant computational overhead as it necessitates training GNNs from scratch and conducting real-time device measurements to assess metrics such as accuracy and efficiency.

To automate model design and optimization, neural architecture search (NAS) was introduced as an AutoML technique, first applied to GNNs design by GraphNAS~\cite{ijcai2020p195}.
However, their focus on optimizing only accuracy fails to address the needs of real-time applications.
Recent efforts have addressed this by introducing hardware-aware NAS methods that aim to optimize both accuracy and efficiency. 
For instance, G-CoS~\cite{zhang2021g} leverage a hardware estimation approach within a GNN-Accelerator co-search framework to balance hardware efficiency and model accuracy for customized hardware platforms. 
Additionally, MaGNAS~\cite{odema2023magnas} utilizes a lookup table (LUT) method to simultaneously search for optimal GNN architectures and mapping schemes for the MPSoC platform. 
However, these hardware-aware NAS solutions are typically tailored for specific hardware designs and may not be easily adaptable to general-purpose computing platforms.
Conversely, HGNAS uses a predictor approach that can be extended to various platforms, including hardware accelerators such as FlowGNN~\cite{sarkar2023flowgnn} and GCNAX~\cite{li2021gcnax}. This extension is made possible by incremental data collection and training for the predictor.

On the other hand, some studies have explored optimization techniques to handle large-scale graph training problems on distributed edge platforms.
For example, SUGAR~\cite{xue2023sugar} leverages graph partitioning and subgraph-level training to enhance the training efficiency of GNNs on large-scale graphs, achieving notable results. 
This method primarily targets the optimization of training processes rather than the design of hardware-efficient architectures. 
Unlike SUGAR, this paper focuses on exploring methods to automatically design efficient GNN models for real-time inference, specifically for edge devices. 
More specifically, we aim to design GNNs that are not only accurate but also tailored to the limited computational resources typically available on edge devices.
\section{Conclusions}\label{sec:conclusions}

In this work, we propose HGNAS, the first hardware-aware framework to automatically explore efficient GNNs for resource-constrained edge devices.
Our objective is to efficiently search the GNN architecture with the optimal hardware efficiency on target platforms, while satisfying the task accuracy.
Specifically, HGNAS adopts a fine-grained design space to facilitate the exploration of high-performance architectures.
Additionally, HGNAS integrates a novel GNN hardware performance prediction method to perceive the hardware efficiency of candidate architectures.
To further streamline the exploration process, HGNAS incorporates the multi-stage hierarchical search strategy, which reduces a single exploration to a few GPU hours.
Extensive experiments show that architectures generated by HGNAS consistently outperform SOTA GNNs, achieving about $10.6\times$ speedup and $82.5\%$ peak memory reduction across edge devices.
We believe that HGNAS has made pivotal progress in bringing GNNs to real-life edge applications.

\bibliographystyle{IEEEtran}
\footnotesize
\begingroup
\bibliography{ref.bib}
\endgroup

\vspace{-10mm}

\begin{IEEEbiography}[{\includegraphics[width=1in,height=1.25in,clip,keepaspectratio]{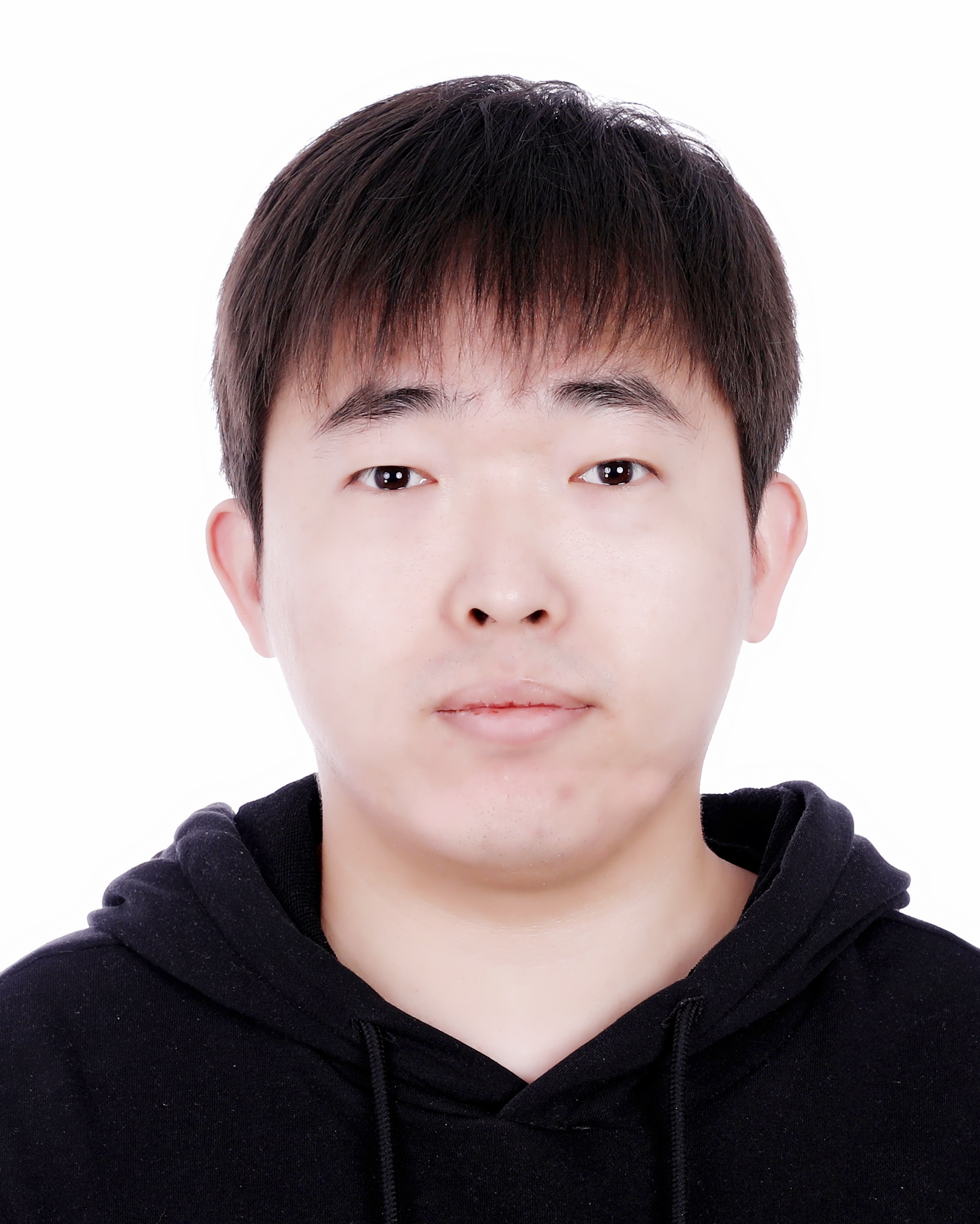}}]{Ao Zhou}

received the B.S. and M.S. degree in software engineering from Beijing University of Technology, Beijing, China, in 2018 and 2021, respectively. He is currently pursuing the Ph.D. degree at the School of Computer Science and Engineering, Beihang University, China. His research interests include GNN acceleration, computer architecture, FPGA accelerator, and heterogeneous computing. He is one of the contributors to the popular graph neural network framework PyTorch Geometric (PyG).

\end{IEEEbiography}

\vspace{-10mm}

\begin{IEEEbiography}[{\includegraphics[width=1in,height=1.25in,clip,keepaspectratio]{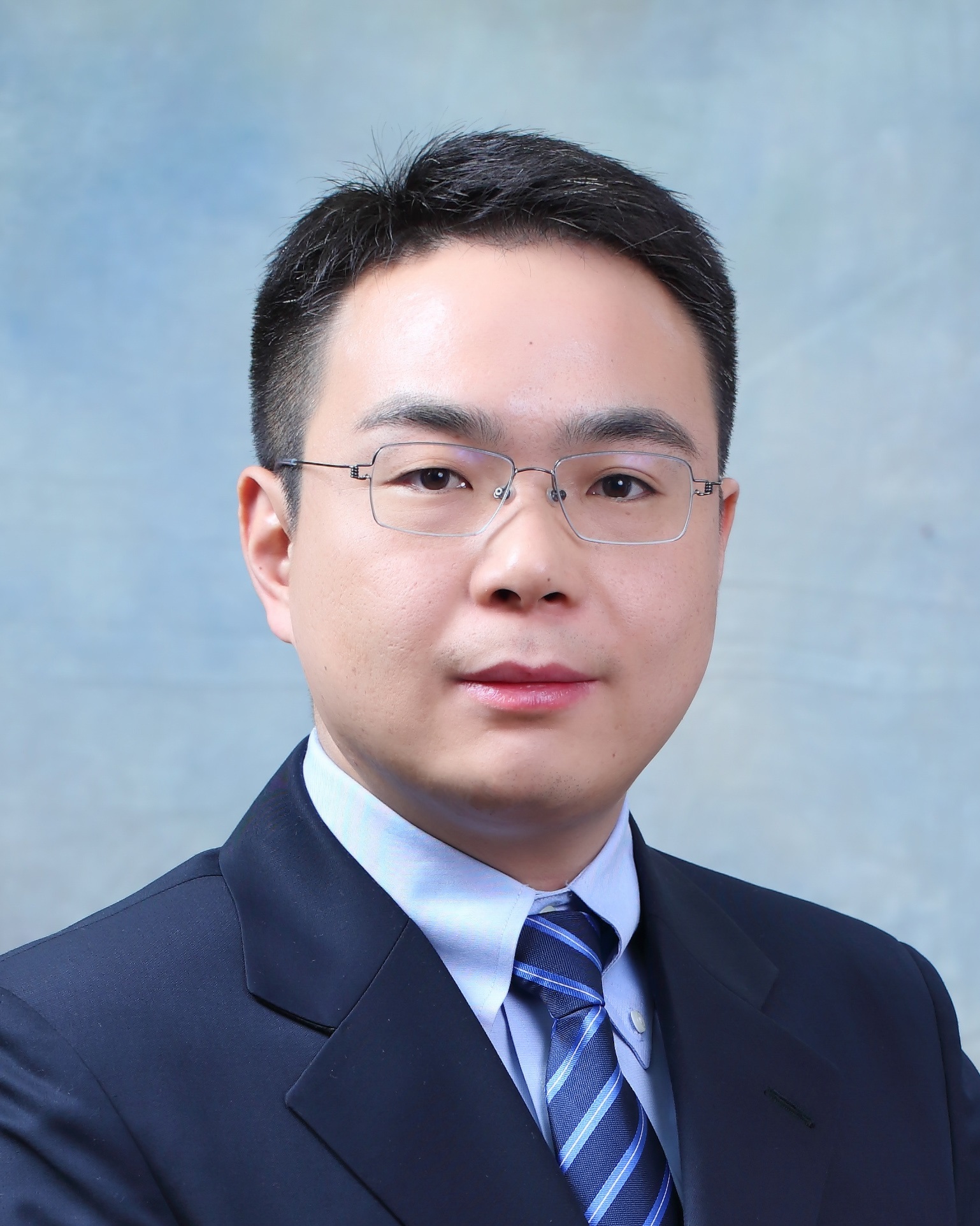}}]{Jianlei Yang}

(S'11-M'14-SM'20) received the B.S. degree in microelectronics from Xidian University, Xi'an, China, in 2009, and the Ph.D. degree in computer science and technology from Tsinghua University, Beijing, China, in 2014.

He is currently an Associate Professor in Beihang University, Beijing, China, with the School of Computer Science and Engineering. From 2014 to 2016, he was a post-doctoral researcher with the Department of ECE, University of Pittsburgh, Pennsylvania, USA.
His current research interests include deep learning accelerators and neuromorphic computing systems.

Dr. Yang was the recipient of the First/Second place on ACM TAU Power Grid Simulation Contest in 2011/2012. He was a recipient of IEEE ICCD Best Paper Award in 2013, ACM GLSVLSI Best Paper Nomination in 2015, IEEE ICESS Best Paper Award in 2017, ACM SIGKDD Best Student Paper Award in 2020.

\end{IEEEbiography}

\begin{IEEEbiography}[{\includegraphics[width=1in,height=1.25in,clip,keepaspectratio]{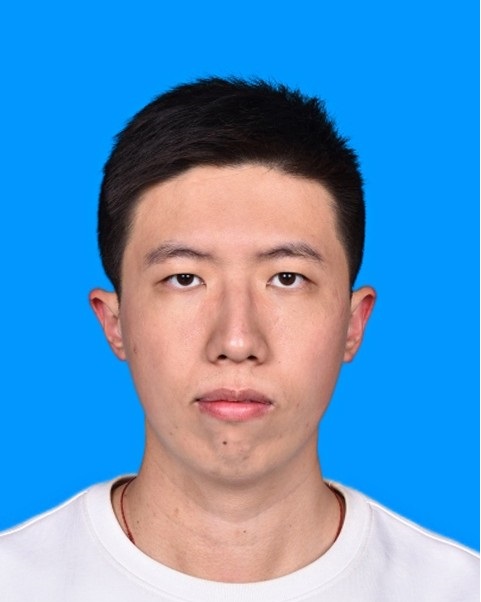}}]{Yingjie Qi}

received the B.S. degree in computer science and technology from Beihang University, Beijing, China, in 2020. He is currently pursuing the Ph.D. degree at the School of Computer Science and Engineering, Beihang University, China. His research interests include graph neural networks acceleration, processing-in-memory architectures and deep learning compilers.

\end{IEEEbiography}

\vspace{-12mm}

\begin{IEEEbiography}[{\includegraphics[width=1in,height=1.25in,clip,keepaspectratio]{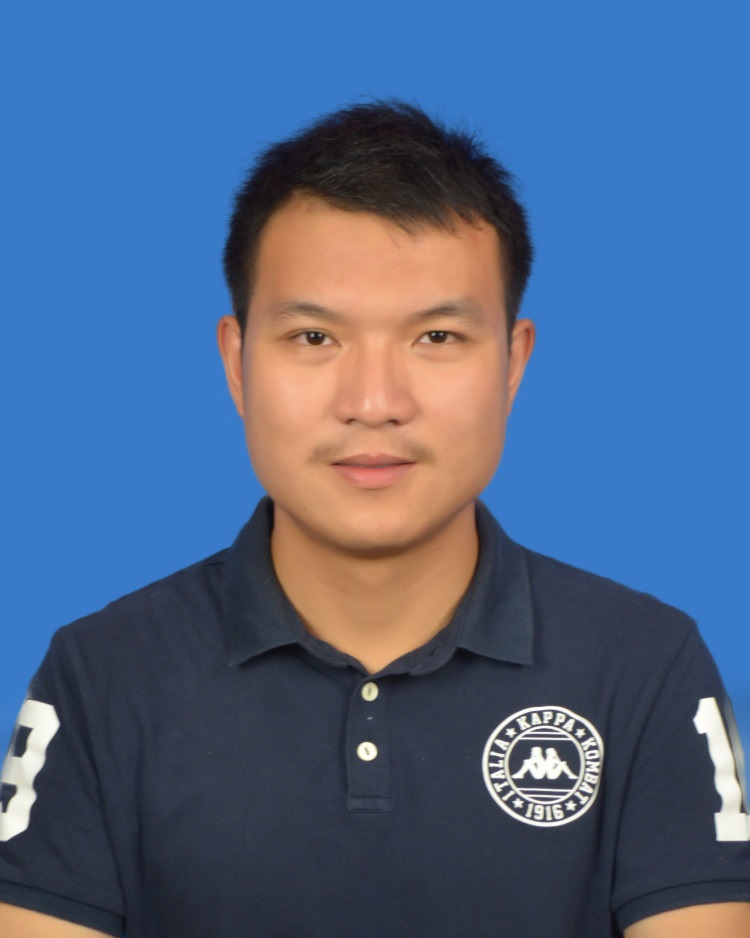}}]{Tong Qiao}

received the B.S. degree in computer science and technology from Beihang University, Beijing, China, in 2020. He is currently pursuing the Ph.D. degree at the School of Computer Science and Engineering, Beihang University, China. His research interests include graph neural networks acceleration, and system for machine learning.

\end{IEEEbiography}

\vspace{-10mm}

\begin{IEEEbiography}[{\includegraphics[width=1in,height=1.25in,clip,keepaspectratio]{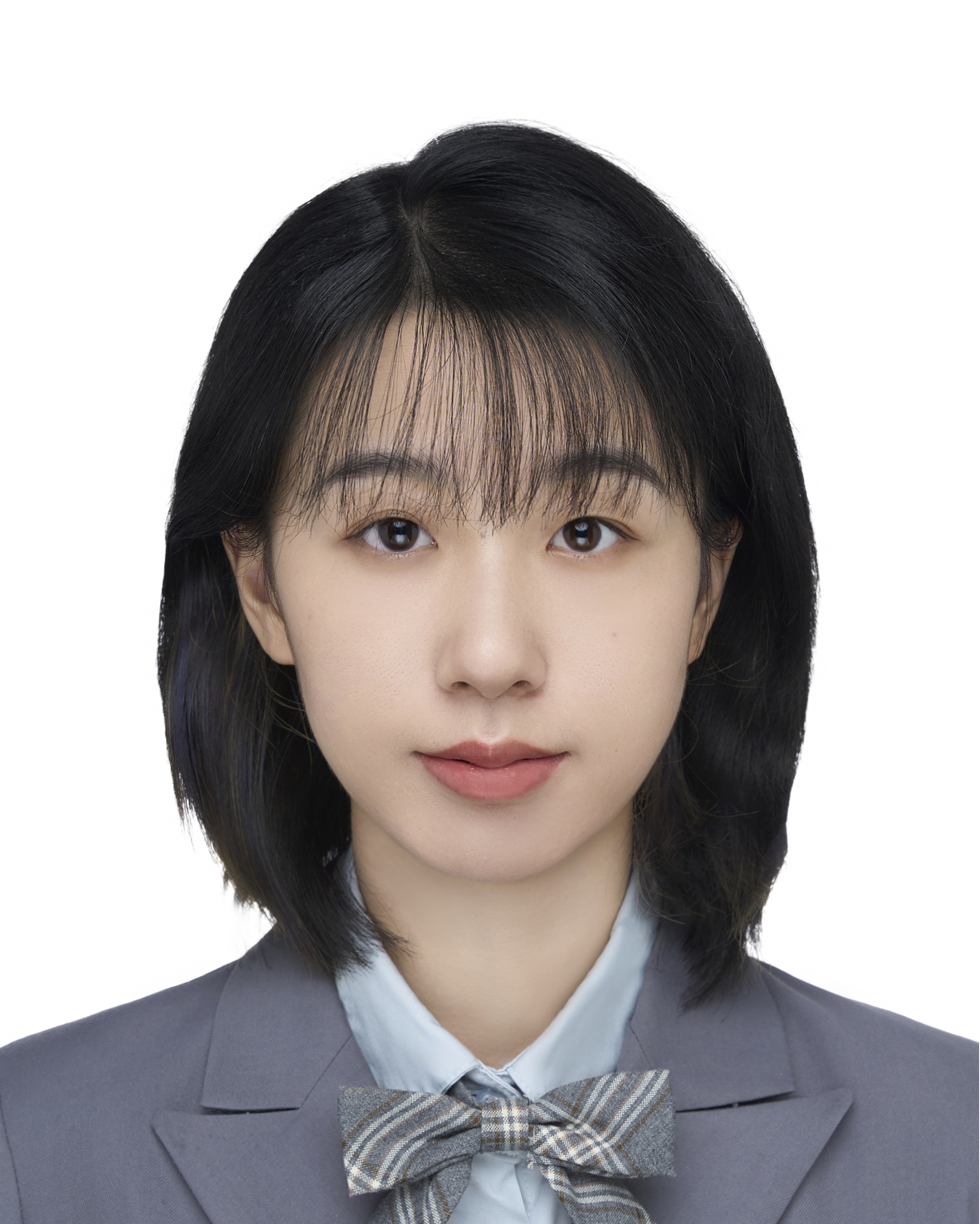}}]{Yumeng Shi}

received the B.S. degree in computer science and technology from Beihang University, Beijing, China, in 2021. She is currently working toward the M.S. degree at the School of Computer Science and Engineering, Beihang University, Beijing, China. Her current research interests include deep learning model compression and hardware/software co-design.

\end{IEEEbiography}

\vspace{-10mm}

\begin{IEEEbiography}[{\includegraphics[width=1in,height=1.25in,clip,keepaspectratio]{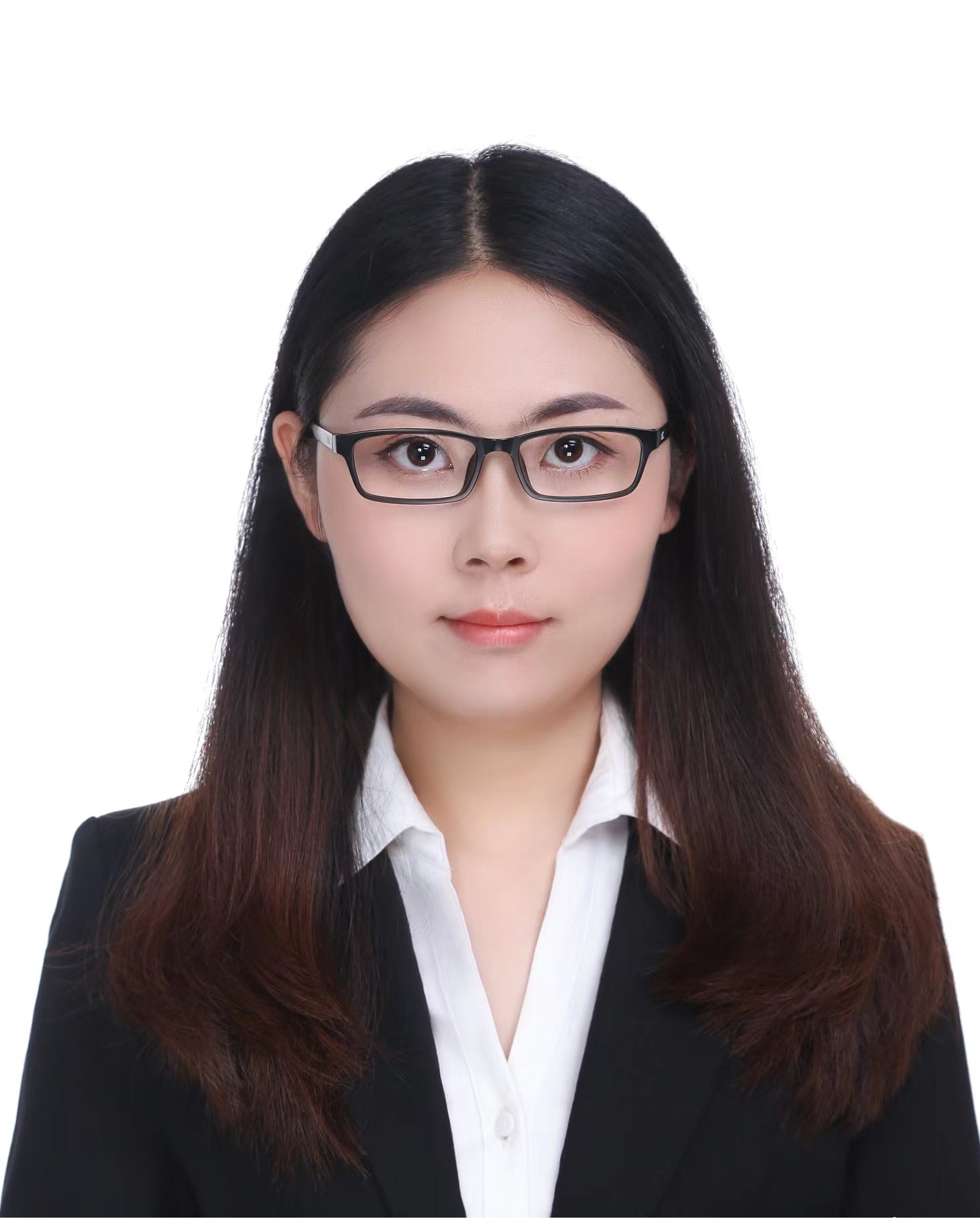}}]{Cenlin Duan}

received the B.S. degree in electronic science and technology from University of Electronic Science and Technology of China, Chengdu, China, in 2015, and the M.S. degree in software engineering from Xidian University, Xi'an, China, in 2018. She is currently pursuing the Ph.D. degree at the School of Integrated Circuit Science and Engineering, Beihang University, Beijing, China. Her current research interests include processing-in-memory architectures and deep learning accelerators.

\end{IEEEbiography}

\vspace{-10mm}
\begin{IEEEbiography}[{\includegraphics[width=1in,height=1.25in,clip,keepaspectratio]{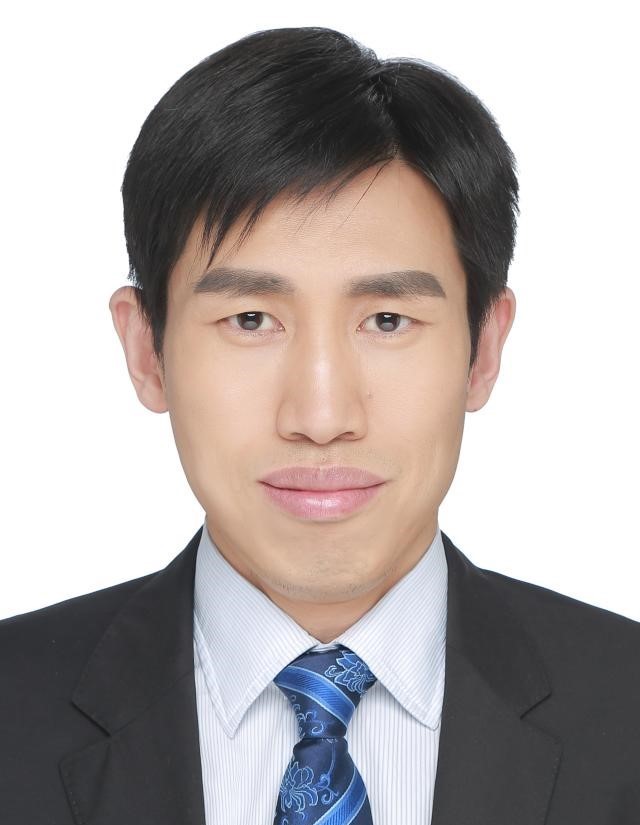}}]{Weisheng Zhao}

(Fellow, IEEE) received the Ph.D. degree in physics from the University of Paris Sud, Paris, France, in 2007.

He is currently a Professor with the School of Integrated Circuit Science and Engineering, Beihang University, Beijing, China. In 2009, he joined the French National Research Center, Paris, as a Tenured Research Scientist. Since 2014, he has been a Distinguished Professor with Beihang University. He has published more than 200 scientific articles in leading journals and conferences, such as \textit{Nature
Electronics}, \textit{Nature Communications}, \textit{Advanced Materials}, IEEE Transactions, ISCA, and DAC. His current research interests include the hybrid integration of nanodevices with CMOS circuit and new nonvolatile memory (40-nm technology node and below) like MRAM circuit and architecture design.

Prof. Zhao was the Editor-in-Chief for the {\sc{IEEE Transactions on Circuits and System I: Regular Paper}} from 2020 to 2023.

\end{IEEEbiography}

\vspace{-10mm}
\begin{IEEEbiography}[{\includegraphics[width=1in,height=1.25in,clip,keepaspectratio]{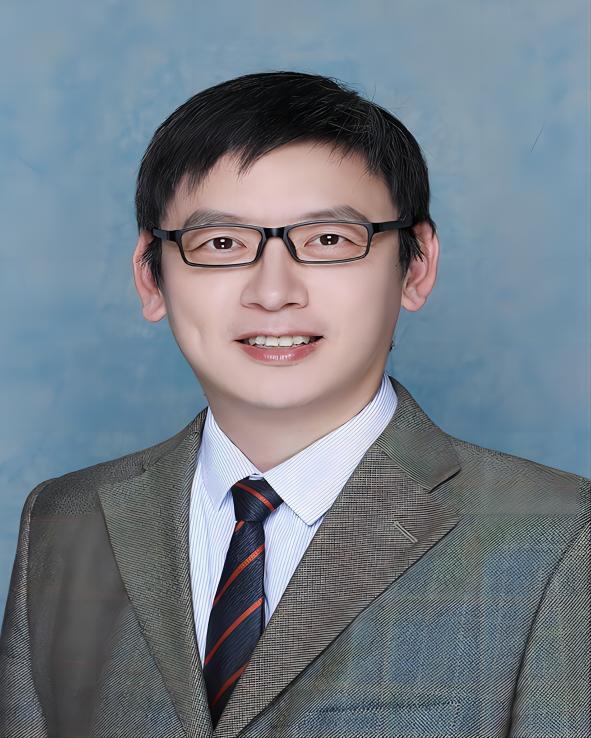}}]{Chunming Hu}

received the PhD degree in computer science and technology from Beihang University, Beijing, China, in 2006.

He is currently a professor and dean of the School of Software, Beihang University, Beijing, China. His current research interests include distributed systems, system virtualization, mobile computing and cloud computing.

Prof. Hu is currently one of the W3C Board of Directors, and serving as the Deputy Director of W3C China.

\end{IEEEbiography}

\end{document}